\documentclass[11pt]{article}

\usepackage[scale=0.75]{geometry}
\usepackage{graphicx}
\usepackage{amsthm}
\usepackage{amsfonts}
\usepackage{amsmath}
\usepackage{mathrsfs}
\usepackage{amssymb}
\usepackage{subcaption}
\usepackage{hyperref}
\usepackage{xcolor}
\usepackage{stmaryrd}

\allowdisplaybreaks
\newcommand{\R}{\mathbb R}

\newcommand{\Q}{\mathcal Q}
\newcommand{\C}{\mathcal C}

\newcommand{\I}{\mathcal I}
\newcommand{\D}{\mathcal D}
\newcommand{\V}{\mathcal V}
\newcommand{\E}{\mathcal E}
\newcommand{\X}{\mathcal X}

\newcommand{\Tmin}{T_\mathrm{min}}
\newcommand{\Tmax}{T_\mathrm{max}}

\newcommand{\Copt}{C_\mathrm{opt}}
\newcommand{\Crelax}{C_\mathrm{relax}}
\newcommand{\Cround}{C_\mathrm{round}}

\newcommand{\deltarelax}{\delta_\mathrm{relax}}
\newcommand{\deltaopt}{\delta_\mathrm{opt}}

\newcommand{\conv}{\mathrm{conv}}

\newcommand{\st}{\text{subject to}}
\newcommand{\mi}{\text{minimize}}

\theoremstyle{definition}

\title{\textbf{Motion Planning around Obstacles \\ with Convex Optimization}}
\author{
Tobia Marcucci$^{1,*}$\and
Mark Petersen$^{2,*}$\and
David von Wrangel$^1$\and
Russ Tedrake$^{1,*}$
}
\date{
$^1$Massachusetts Institute of Technology, Cambridge MA, USA \\
{\small \texttt{\{tobiam,wrangel,russt\}@mit.edu}} \\
$^2$Harvard University, Cambridge MA, USA \\
{\small \texttt{markpetersen@g.harvard.edu}} \\
\footnotetext[1]{Contributed equally to this work.}
}

\begin{document}
\maketitle

\begin{abstract}
Trajectory optimization offers mature tools for motion planning in high-dimensional spaces under dynamic constraints.
However, when facing complex configuration spaces, cluttered with obstacles, roboticists typically fall back to sampling-based planners that struggle in very high dimensions and with continuous differential constraints.
Indeed, obstacles are the source of many textbook examples of problematic nonconvexities in the trajectory-optimization problem.
Here we show that convex optimization can, in fact, be used to reliably plan trajectories around obstacles.
Specifically, we consider planning problems with collision-avoidance constraints, as well as cost penalties and hard constraints on the shape, the duration, and the velocity of the trajectory.
Combining the properties of B\'{e}zier curves with a recently-proposed framework for finding shortest paths in Graphs of Convex Sets (GCS), we formulate the planning problem as a compact mixed-integer optimization.
In stark contrast with existing mixed-integer planners, the convex relaxation of our programs is very tight, and a cheap rounding of its solution is typically sufficient to design globally-optimal trajectories.
This reduces the mixed-integer program back to a simple convex optimization, and automatically provides optimality bounds for the planned trajectories.
We name the proposed planner GCS, after its underlying optimization framework.
We demonstrate GCS in simulation on a variety of robotic platforms, including a quadrotor flying through buildings and a dual-arm manipulator (with fourteen degrees of freedom) moving in a confined space.
Using numerical experiments on a seven-degree-of-freedom manipulator, we show that GCS can outperform widely-used sampling-based planners by finding higher-quality trajectories in less time.
\end{abstract}

\section{Introduction}
In this paper we consider the problem of designing continuous collision-free trajectories for robots moving in environments with obstacles.
A wide array of techniques can be found in the literature to tackle this long-standing problem in robotics~\cite{hoy2015algorithms}, and selecting the right one requires compromising between multiple features of the problem at hand: dimensionality and complexity of the environment, dynamic constraints, computation limits, completeness and optimality requirements.

Methods based on direct trajectory optimization~\cite{diehl2006fast,augugliaro2012generation,schulman2014motion,majumdar2017funnel,zhang2020optimization} can design trajectories in high-dimensional spaces while taking into account the robot kinematics and dynamics.
However, by transcribing the planning problem as a nonconvex program, and by relying on local optimization, these techniques can fail in finding a collision-free trajectory, especially if the robot configuration space is cluttered.
In these scenarios, roboticists typically fall back to sampling-based planners~\cite{kavraki1996probabilistic,lavalle1998rapidly} (see also the more recent review~\cite{elbanhawi2014sampling}).
These algorithms are \emph{probabilistically complete}, meaning that, if a feasible path exists, they will eventually find one, regardless of the complexity of the configuration space~\cite[Chapter~5]{lavalle2006planning}.
This guarantee, however, comes at a cost.
Although many sampling-based planners support ``kinodynamic'' constraints~\cite{webb2013kinodynamic,goretkin2013optimal,wu2020r3t}, continuous differential constraints are difficult to impose on discrete samples, making the kinodynamic versions of the classical sampling-based algorithms much less successful in practice.
In addition, even using asymptotically-optimal sampling-based planners~\cite{karaman2011sampling,gammell2014informed,janson2015fast}, the trajectories we design can be considerably suboptimal in practice, where only a finite number of samples can be taken.
For certain classes of dynamical systems, hybrid approaches, where a trajectory-optimization planner is driven by a higher-level graph search, have been shown to overcome part of these difficulties~\cite{natarajan2021interleaving}.
Still, these multi-layer architectures do not offer a unified formulation of the planning problem as a single optimization problem.

The promise of the planners based on Mixed-Integer Convex Programming (MICP)~\cite{schouwenaars2001mixed,richards2002aircraft,mellinger2012mixed,webb2013kinodynamic,deits2015efficient} is to take the best of the two worlds above: the completeness of sampling-based algorithms, and the ease with which trajectory optimization handles the robot kinematics and dynamics; with the added bonus of global optimality and within a single optimization framework.
The spread of MICP techniques, however, is strongly limited by their runtimes: even for small-scale problems, these methods can require several minutes to design a trajectory.
Only recently, collision-free planners entirely based on convex optimization have been proposed~\cite{el2021piecewise}, but their application is currently limited to purely-geometric path planning in low-dimensional spaces.

In this paper, we focus on a limited but important class of motion-planning problems with differential constraints, and we present a planner that, although being based on MICP, reliably solves very high-dimensional problems in a few seconds, through a single convex program.

\subsection{Contribution}
We consider a formulation of the collision-free planning problem similar to the one from~\cite{deits2015efficient}.
In particular,  we assume the robot configuration space to be partitioned into a collection of ``safe'' convex regions, i.e., regions that do not intersect with any of the obstacles.
In the special case of polygonal obstacles, this partition can be constructed exactly.
More generally, approximate decompositions can be efficiently obtained using existing algorithms~\cite{ayanian10,deits2015computing}, as well as newly-developed techniques tailored to the complex configuration spaces of multi-link kinematic trees~\cite{amice2022finding}.
Our goal is then to design a continuous trajectory that is entirely contained in the union of the safe regions.
The optimality criterion and the additional constraints are allowed to depend on the shape, the duration, and the velocity of the trajectory.

The main technical contribution of this work is showing that the trajectory-design problem just described can be formulated as a shortest-path problem in Graphs of Convex Sets (GCS): a recently-studied class of optimizations that lends itself to very efficient mixed-integer programming~\cite{marcucci2021shortest}.
Existing MICP planners parameterize a single trajectory and use binary variables to assign each of its segments to a safe region.
Conversely, with the proposed planner, which we name GCS, the safe regions are connected through an adjacency graph and are each assigned a trajectory segment.
The optimal probabilities of transitioning between the regions are then computed via an efficient blending of convex and graph optimization.
We show that the MICPs constructed in this way have very tight convex relaxations and, in the great majority of practical cases, a single convex program, together with a cheap rounding step, is sufficient to identify a globally-optimal collision-free trajectory.
Furthermore, by comparing the costs of the convex relaxation and the rounded trajectory, GCS automatically provides a tight bound on the optimality of the motion plan.

To parameterize trajectories we use B\'{e}zier curves: a relatively common tool in motion planning (see e.g.~\cite{flores2008real,lau2009kinodynamic,csomay2022multi}) whose properties are very well suited for mixed-integer programming~\cite{koolen2020balance}.
This parameterization enables simple convex formulations of the collision-avoidance constraints and, when incorporated in our workflow, leads to very tractable convex optimizations; typically Second-Order-Cone Programs (SOCPs).
This is in contrast with existing MICP planners, which require expensive semidefinite constraints to design trajectories that are differentiable more than three times~\cite{deits2015efficient}.
(Note that the requirement of smooth trajectories is of practical nature: to exploit the differential-flatness properties of quadrotors, for example, it is necessary to design trajectories that are differentiable at least four times~\cite{mellinger2011minimum}.)

We demonstrate GCS on a variety of planning problems, ranging from an intricate maze to a quadrotor flying through buildings and a fourteen-dimensional dual-arm manipulation task.
The numerical results show that, besides significantly improving on state-of-the-art MICP planners, our relatively unoptimized implementation of GCS can also outperform widely-used sampling-based planners by finding higher-quality trajectories in lower, and consistent, runtimes.

\section{Problem Statement}
\label{sec:problem_statement}
In this section we state the motion-planning problem addressed in this paper in abstract terms, as an optimization over the infinite-dimensional space of trajectories.
It will be the goal of Section~\ref{sec:collision_free_motion_planning_as_spp} to present our finite-dimensional transcription of this optimization, which will then be tackled using practical convex programming.

As in~\cite{deits2015efficient}, we look at the problem of planning around obstacles as the problem of navigating within a collection of ``safe'' regions.
More precisely, we assume the set $\Q \subset \R^n$ of collision-free robot configurations to be decomposed into a family of (possibly overlapping) bounded convex sets $\Q_i \subseteq \Q$, with $i$ in a finite index set $\I$.
For polyhedral obstacles this decomposition can be exact, i.e. $\bigcup_{i \in \I} \Q_i = \Q$, while more complex configuration spaces can be decomposed approximately using efficient existing algorithms~\cite{deits2015computing,amice2022finding}.
Given the regions $\Q_i$, our goal is to find a time $T \in \R_{> 0}$ and a trajectory $q : [0, T] \rightarrow \Q$ that are a solution of the following optimization problem:\footnote{
In Section~\ref{sec:additional_differential_costs_and_constraints} we show how penalties on the second and higher derivatives of $q$ can be approximately integrated in our problem formulation.
Further costs and constraints are discussed in Section~\ref{sec:additional_costs_and_constraints}.
}
\begin{subequations}
\label{eq:statement}
\begin{align}
\label{eq:statement_objective}
\mi \quad & a T + b L(q, T) + c E(\dot q, T) \\
\st \quad
\label{eq:statement_continuity}
& q \in \C^\eta, \\
\label{eq:statement_disjuctive}
& q(t) \in \bigcup_{i \in \I} \Q_i, && \forall t \in [0,T], \\
\label{eq:statement_derivative_constraints}
& \dot q(t) \in \D, && \forall t \in [0,T], \\
\label{eq:statement_duration}
& T \in [\Tmin, \Tmax], \\
\label{eq:statement_boundary_conditions_0}
& q(0) = q_0, \ q(T) = q_T, \\
\label{eq:statement_boundary_conditions_1}
& \dot q(0) = \dot q_0, \ \dot q(T) = \dot q_T.
\end{align}
\end{subequations}
The objective is a weighted sum, with user-specified weights $a,b,c \in \R_{\geq 0}$, of the trajectory duration $T$, the length $L (q,T)$ of the trajectory, and the energy $E(\dot q, T)$ of the time derivative of the trajectory.
Specifically, the latter two quantities are defined as
\begin{align}
\label{eq:L_and_E}
L(q, T) := \int_0^T \|\dot q(t)\|_2 dt
\quad \text{and} \quad
E(\dot q, T) := \int_0^T \|\dot q(t)\|_2^2 dt.
\end{align}
Constraint~\eqref{eq:statement_continuity} asks the trajectory to be continuously differentiable $\eta$ times.
Constraint~\eqref{eq:statement_disjuctive} ensures that $q$ is contained in the safe sets, and hence is collision free at all times.
(Note that this is a stronger constraint than is usual in sampling-based motion planning, where trajectories are typically checked to be collision-free only at a finite number of points.)
The set $\D$ in~\eqref{eq:statement_derivative_constraints} is required to be convex and can be used to enforce hard limits on the robot velocity.
The bounds on the trajectory duration in~\eqref{eq:statement_duration} are such that $\Tmax  \geq \Tmin > 0$.
Finally, the constraints~\eqref{eq:statement_boundary_conditions_0} and~\eqref{eq:statement_boundary_conditions_1} enforce the boundary conditions on $q$ and its time derivative.

The coupling between the trajectory $q$ and its duration $T$ makes it hard to work with problem~\eqref{eq:statement} directly.
Similarly to~\cite{verscheure2009time}, we break this coupling by introducing the path coordinate $s \in [0,S]$, where $S$ has fixed positive value.
We relate the coordinate $s$ to the time variable $t$ via the scaling function $t = h(s)$: the map $h$ is required to be  monotonically increasing, and such that $h(0)=0$ and $h(S) = T$.
Expressing the trajectory $q$ as a function of $s$, we get the curve $r(s) := q(h(s))$.
Through a few simple manipulations, we restate problem~\eqref{eq:statement} in terms of the decision variables $r$ and $h$ as
\begin{subequations}
\label{eq:scaled}
\begin{align}
\label{eq:scaled_objective}
\mi \quad & a h(S) + b L(r, S) + c E \left(\dot r / \sqrt{\dot h}, S\right) \\
\st \quad
\label{eq:scaled_continuity}
& r \circ h^{-1} \in \C^\eta, \\
\label{eq:scaled_disjuctive}
& r(s) \in \bigcup_{i \in \I} \Q_i, && \forall s \in [0,S], \\
\label{eq:scaled_velocity}
& \dot r(s) \in \dot h(s) \D, \ \dot h(s) > 0, && \forall s \in [0,S], \\
\label{eq:scaled_duration}
& h(0) = 0, \ h(S) \in [\Tmin, \Tmax], \\
\label{eq:scaled_boundary_conditions_0}
& r(0) = q_0, \ r(S) = q_T, \\
\label{eq:scaled_boundary_conditions_1}
& \dot r(0) = \dot h(0) \dot q_0, \ \dot r(S) = \dot h(S) \dot q_T.
\end{align}
\end{subequations}
In particular, we have used the chain rule to substitute $\dot q (t)$ with $\dot r(s) / \dot h(s)$, and we have changed integration variable in~\eqref{eq:L_and_E} from $t$ to $s$.
This makes $\dot r / \sqrt{\dot h} : [0,S] \rightarrow \R^n$ the argument of the energy function in the objective.
The symbol $\circ$ in~\eqref{eq:scaled_continuity} denotes the composition operator: notice that the function $h$ is guaranteed to be invertible by the positivity of $\dot h$ from~\eqref{eq:scaled_velocity}.
Finally, again by the chain rule, the right-hand sides of the velocity constraints in~\eqref{eq:scaled_velocity} and~\eqref{eq:scaled_boundary_conditions_1} are multiplied by the derivative $\dot h$ of the time scaling.


\section{Background on B\'{e}zier Curves}
\label{sec:background_on_bezier_curves}
In order to tackle problem~\eqref{eq:scaled} numerically, it is necessary to parameterize the functions $r$ and $h$ through a finite number of decision variables.
To this end, in Section~\ref{sec:collision_free_motion_planning_as_spp}, we will employ B\'{e}zier curves.
The goal of this section is to recall the definition and the basic properties of this family of curves.

A B\'{e}zier curve is constructed using Bernstein polynomials.
The $k$th Bernstein polynomial of degree $d$, with $k =0, \ldots, d$, is defined as
$$
\beta_{k,d} (s) := \binom{d}{k} s^k (1 - s)^{d-k},
$$
where $s \in [0,1]$.
Note that the Bernstein polynomials of degree $d$ are nonnegative and, by the binomial theorem, they sum up to one.
Therefore, for each fixed $s \in [0,1]$, the scalars $\{\beta_{k,d}(s)\}_{k=0}^d$ can be thought of as the coefficients of a convex combination.
B\'{e}zier curves are obtained using these coefficients to combine a given set of $d+1$ \emph{control points} $\gamma_k \in \R^n$:
$$
\gamma(s) := \sum_{k = 0}^d \beta_{k,d}(s) \gamma_k.
$$

It is easily verified that B\'{e}zier curves enjoy the following properties.
\begin{itemize}
\item \emph{Endpoint values.}
The curve $\gamma$ starts at the first control point and ends at the last control point: $\gamma(0) = \gamma_0$ and $\gamma(1) = \gamma_d$.
\item \emph{Convex hull.}
The curve $\gamma$ is entirely contained in the convex hull of its control points: $\gamma(s) \in \conv(\{\gamma_k\}_{k=0}^d)$ for all $s \in [0,1]$.
\item \emph{Derivative.}
The derivative $\dot \gamma$ of the curve $\gamma$ is a B\'{e}zier curve of degree $d-1$ with control points $\dot \gamma_k = d (\gamma_{k+1} - \gamma_k)$ for $k = 0, \ldots, d-1$.
\item \emph{Integral of convex function.}
For a convex function $f:\R^n \rightarrow \R$, we have\footnote{
To prove~\eqref{eq:integral_convex}, one uses the convexity of $f$, which gives $f(\gamma(s)) \leq \sum_{k=0}^d \beta_{k,d}(s)f(\gamma_k)$, and the formula $\int_0^1 \beta_{k,d} (s) ds = 1/(d+1)$ for the integration of Bernstein polynomials.}
\begin{align}
\label{eq:integral_convex}
\int_0^1 f(\gamma(s)) ds \leq \frac{1}{d+1} \sum_{k=0}^d f(\gamma_k).
\end{align}
\end{itemize}

\section{The Optimization Framework}
\label{sec:the_optimization_framework}
Our strategy for solving problem~\eqref{eq:scaled} is to first transcribe it as a Shortest-Path Problem (SPP) in GCS, and then use the techniques recently presented in~\cite{marcucci2021shortest} to formulate this SPP as a compact MICP.
As we will see in Section~\ref{sec:numerical_results}, the convex relaxation of this MICP is extremely tight in practice, up to the point that a cheap rounding of its solution is almost always sufficient to design a globally-optimal trajectory.
In this section, we give a formal statement of the SPP in GCS and we propose a simple randomized rounding for the convex relaxation of our MICP.
The latter will effectively reduce the computational cost of the MICP to that of a convex program.

\subsection{Shortest Paths in Graphs of Convex Sets}
\label{sec:finding_shortest_paths_in_gcs}
The Shortest-Path Problem (SPP) in GCS generalizes the classical SPP with nonnegative edge lengths.
We are given a directed graph $G := (\V, \E)$ with vertex set $\V$ and edge set $\E \subset \V^2$.
Each vertex $v \in \V$ is paired with a bounded convex set $\X_v$, and a point $x_v$ contained in it.
In contrast to the classical SPP, where edge lengths are fixed scalars, here the length of an edge $e=(u,v)$ is determined by the continuous values of $x_u$ and $x_v$ via the expression $\ell_e(x_u,x_v)$.
The function $\ell_e$ is assumed to be convex and to take nonnegative values.
Convex constraints of the form $(x_u, x_v) \in \X_e$ are allowed to couple the endpoints of edge $e:=(u,v)$.
A path $p$ in the graph $G$ is defined as a sequence of distinct vertices that connects the source vertex $\sigma \in \V$ to the target vertex $\tau \in \V$.
Denoting with $\E_p$ the set of edges traversed by the path $p$, and with $\mathcal P$ the family of all $\sigma$-$\tau$ paths in the graph $G$, the SPP in graphs of convex sets is stated as
\begin{subequations}
\label{eq:spp}
\begin{align}
\label{eq:spp_objective}
\mi \quad & \sum_{e := (u,v) \in \E_p} \ell_e(x_u,x_v) \\
\st \quad
\label{eq:spp_path}
& p \in \mathcal P, \\
\label{eq:spp_v}
& x_v \in \X_v, && \forall v \in p, \\
\label{eq:spp_e}
& (x_u, x_v) \in \X_e, && \forall e:=(u,v) \in \E_p.
\end{align}
\end{subequations}
Here the decision variables are the discrete path $p$ and the continuous values $x_v$.
The objective~\eqref{eq:spp_objective} minimizes the length of the path $p$, defined as the sum of the lengths of its edges.
Constraint~\eqref{eq:spp_path} asks $p$ to be a valid path connecting $\sigma$ to $\tau$.
Importantly, the convex conditions~\eqref{eq:spp_v} and~\eqref{eq:spp_e} constrain only the continuous variables paired with the vertices visited by the path $p$, and do not apply to the remaining vertices in the graph.
Unlike the classical SPP with nonnegative edge lengths, which is easily solvable in polynomial time, the SPP in GCS can be verified to be NP-hard~\cite[Theorem~1]{marcucci2021shortest}.

\subsection{Rounding the Convex Relaxation of the Shortest-Path Problem}
\label{sec:rounding_the_convex_relaxation_of_the_spp}
Using recently-developed techniques, problem~\eqref{eq:spp} is formulated as a compact MICP with very tight convex relaxation~\cite[Equation~21]{marcucci2021shortest}.
In this paper, instead of tackling this MICP with an exact branch-and-bound algorithm, we solve its convex relaxation and we recover an approximate solution via a cheap randomized rounding, that is tailored to the graph structure beneath problem~\eqref{eq:spp}.
Given the hardness of~\eqref{eq:spp}, this approach cannot be guaranteed to work for all instances.
Nevertheless, for our planning problems, this strategy turns out to be extremely effective in practice.
In addition, this workflow automatically provides us with a bound on the optimality of the  approximate solution we identify.
In fact, denoting with $\Crelax$ the cost of the convex relaxation, with $\Copt$ the optimal value of~\eqref{eq:spp}, and with $\Cround$ the cost of the rounded solution, we have $\Crelax \leq \Copt \leq \Cround$.
The optimality gap of the rounded solution $\deltaopt := (\Cround - \Copt)/\Copt$ can be then overestimated as $\deltarelax := (\Cround - \Crelax)/\Crelax$ with no additional computation.

For the rounding step we propose a randomized strategy.
The MICP from~\cite{marcucci2021shortest} parameterizes a path $p$ by using a binary variable $\varphi_e$ per edge $e \in \E$, with $\varphi_e = 1$ if and only if $e \in \E_p$.
In the convex relaxation, the binary requirement is relaxed to $\varphi_e \in [0,1]$ and the optimal value of $\varphi_e$ is naturally interpreted as the probability of the edge $e$ being a part of the shortest path.
To round these probabilities we then run a randomized depth-first search with backtracking.
We initialize our candidate path as $p := (\sigma)$, and we denote with $\E_u$ the set of edges $e := (u,v)$ that connect $u$ to a vertex $v$ that the rounding algorithm has not visited yet.
At each iteration, calling $u$ the last vertex in the path $p$, we traverse the edge $e := (u,v) \in \E_u$ with probability $\varphi_e / \sum_{e' \in \E_u} \varphi_{e'}$, and we append a new vertex $v$ to the path $p$.
If a dead end occurs, i.e. if $\varphi_e = 0$ for all $e \in \E_u$, we backtrack to the last vertex in $p$ that admits a way out.
The algorithm terminates when $v = \tau$ and the target is reached.\footnote{
Making this rounding strategy deterministic by, e.g., selecting at each iteration the edge $e \in \E_u$ with larger probability $\varphi_e$ is, in general, a bad idea.
To see this, imagine a graph where multiple paths represent the same underlying decision (e.g. multiple symmetrical solutions).
Since the convex relaxation will equally split the probability of this decision being optimal between the edges of these many paths, a greedy deterministic search might end up selecting an alternative path, corresponding to a decision that is overall less likely to be optimal.
Conversely, in the same scenario, a randomized rounding correctly weights the two decisions (in expectation).
}
Once a path $p$ is identified, its cost, together with the optimal values of the continuous variables $x_v$, is recovered by solving a small convex program:
\begin{align}
\label{eq:path_evaluation}
\mi~\eqref{eq:spp_objective} \ \st \ \eqref{eq:spp_v} \ \text{and} \ \eqref{eq:spp_e}.
\end{align}
It is easily verified that this rounding strategy always finds a valid path (provided that the convex relaxation of the MICP is feasible).
On the other hand, the cost of the path $p$ we find can in principle be infinite, since there might not be an assignment for the continuous variables $\{x_v\}_{v \in p}$ that satisfies the constraints~\eqref{eq:spp_v} and~\eqref{eq:spp_e}.

To increase our chances of finding a high-quality approximate solution, we apply the randomized rounding multiple times.
First we run the depth-first search until $N$ distinct paths are identified, or a maximum number $M$ of trials is reached.
Then we evaluate the cost of each distinct path by solving a convex program of the form~\eqref{eq:path_evaluation}, and we return the rounded solution of lowest cost $\Cround$.\footnote{
This sequence of convex optimizations is stopped early if the cost of a path coincides with the cost of the convex relaxation $\Crelax$, since this proves the global optimality of the path at hand.
}
We emphasize that this process is extremely cheap: the runtime of a depth-first search is practically zero (since it is a purely-discrete search in the graph $G$), while the convex programs~\eqref{eq:path_evaluation} are tiny, very sparse, and parallelizable.
In this paper we set $N := 10$ and $M := 100$.
These values lead to rounding times that are negligible with respect to the solution time of the convex relaxation and, in our experiments, they are typically sufficient to solve the planning problem to global optimality.

Many more details on the MICP formulation of~\eqref{eq:spp} and its convex relaxation can be found in~\cite{marcucci2021shortest}.
For the scope of this paper, we will treat the framework from~\cite{marcucci2021shortest} as a modeling language that allows us to formulate, and efficiently solve, an SPP in GCS just by providing the graph $G$, the edge lengths $\ell_e$, and the sets $\X_v$ and $\X_e$.


\section{Collision-Free Motion Planning using Graphs of Convex Sets}
\label{sec:collision_free_motion_planning_as_spp}
We now illustrate how problem~\eqref{eq:scaled} can be transcribed as an SPP in GCS.
As seen in the previous section, to formulate an SPP in GCS we need to: define a graph $G := (\V, \E)$, assign a set $\X_v$ to each vertex $v \in \V$, and pair each edge $e \in \E$ with a constraint set $\X_e$ and a length function $\ell_e$.
Below we describe how each of these components is constructed.
At a high level, the plan is to pair each safe region $\Q_i$ with two B\'{e}zier curves: a trajectory segment $r_i$, and a time-scaling function $h_i$ that dictates the speed at which the curve $r_i$ is traveled.
The functions $r$ and $h$ in problem~\eqref{eq:scaled} will be then reconstructed by sequencing the B\'{e}zier curves $r_i$ and $h_i$ paired with the regions $\Q_i$ that are selected by the SPP.
Figure~\ref{fig:spp_construction} provides a visual support to the upcoming discussion.

\subsection{The Graph $G$}
\label{sec:the_graph_G}
We let the vertex set $\V$ contain a vertex $i$ per safe set $\Q_i$ in the decomposition of the configuration space.\footnote{As we will see in Section~\ref{sec:the_convex_sets_Xv}, the safe set $\Q_i$ does not coincide with the convex set $\X_i$ paired with vertex $i$ in the SPP.}
In addition, we introduce a source vertex $\sigma$ and a target vertex $\tau$: these will be used to enforce the boundary conditions~\eqref{eq:scaled_duration}--\eqref{eq:scaled_boundary_conditions_1}.
Overall, we then have $\V := \I \cup \{\sigma, \tau\}$.

\begin{figure}[t]
\centering
\begin{subfigure}[b]{0.24\textwidth}
\centering
\includegraphics[width=\textwidth]{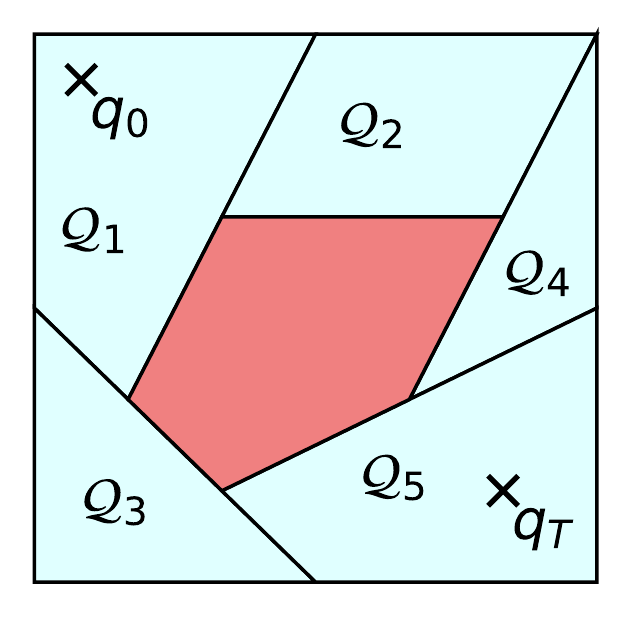}
\caption{}
\label{fig:safe_sets}
\end{subfigure}
\hfill
\begin{subfigure}[b]{0.166\textwidth}
\centering
\includegraphics[width=\textwidth]{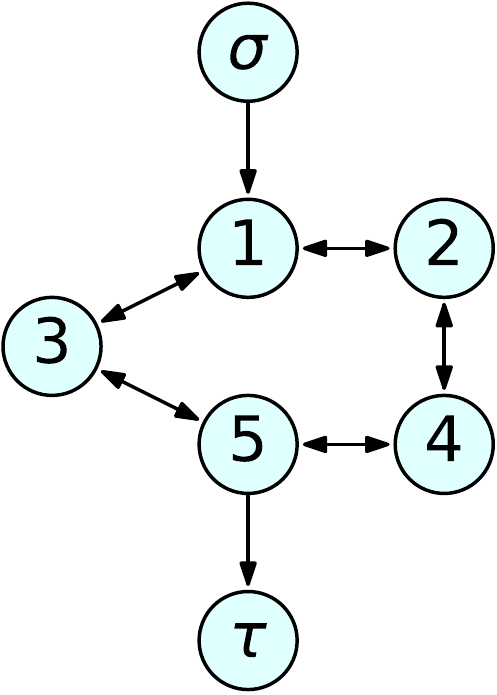}
\caption{}
\label{fig:edge_set}
\end{subfigure}
\hfill
\begin{subfigure}[b]{0.24\textwidth}
\centering
\includegraphics[width=\textwidth]{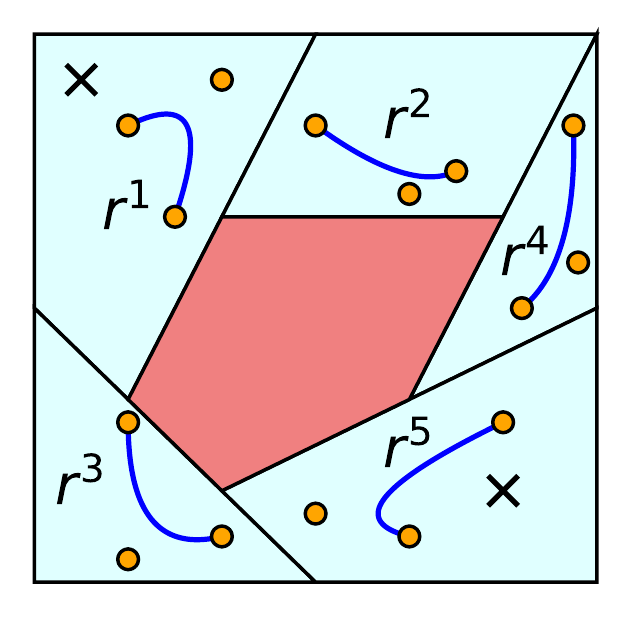}
\caption{}
\label{fig:bezier}
\end{subfigure}
\hfill
\begin{subfigure}[b]{0.24\textwidth}
\centering
\includegraphics[width=\textwidth]{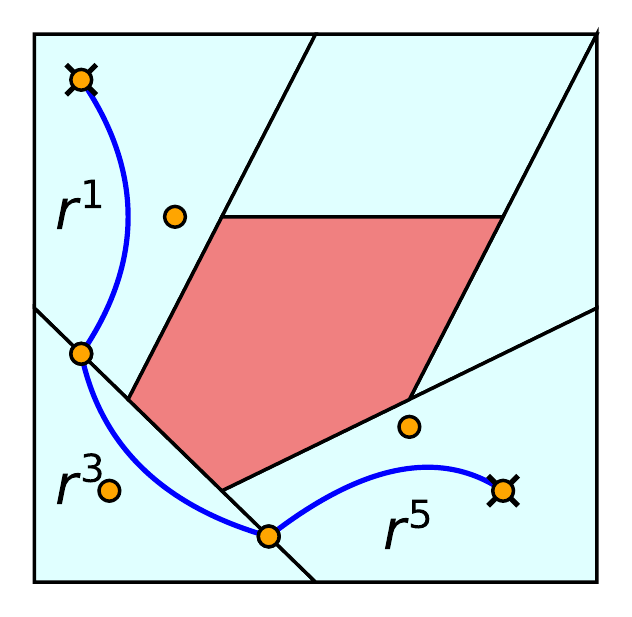}
\caption{}
\label{fig:continuous_bezier}
\end{subfigure}
\caption{Formulation of the collision-free motion-planning problem as an SPP in GCS.
The red region is the obstacle to be avoided, the light-blue regions $\Q_i$ partition the free space.
(a)~The collision-free regions $\Q_i$, the starting point $q_0$, and the ending point $q_T$.
(b)~The graph $G$ obtained by connecting intersecting regions, with the source vertex $\sigma$ and the target vertex $\tau$ added to enforce the initial and terminal conditions, respectively.
(c)~The B\'{e}zier curves $r^i$ associated with each region (curves in blue, control points $r_{i,k}$ in orange).
(d)~A continuous collision-free trajectory $r$ corresponding to the path $p := (\sigma,1,3,5,\tau)$.
}
\label{fig:spp_construction}
\end{figure}

We include in the edge set $\E$ all the edges $(i,j)$ such that the intersection of $\Q_i$ and $\Q_j$ is nonempty.
Note that, by the symmetry of this condition, $(i,j) \in \E$ implies $(j,i) \in \E$.
Similarly, we let $(\sigma,i) \in \E$ and $(i,\tau) \in \E$ if the set $\Q_i$ contains the points $q_0$ and $q_T$, respectively.
In symbols,
$$
\E := \{(i,j): \Q_i \cap \Q_j \neq \emptyset \} \cup \{(\sigma, i): q_0 \in \Q_i \} \cup \{(i,\tau): q_T \in \Q_i\}.
$$
Figure~\ref{fig:edge_set} shows the graph corresponding to the collision-free regions $\Q_i$, the staring point $q_0$, and the ending point $q_T$ depicted in Figure~\ref{fig:safe_sets}.

\subsection{The Convex Sets $\X_v$}
\label{sec:the_convex_sets_Xv}
The source $\sigma$ and the target $\tau$ are auxiliary vertices used to enforce the boundary conditions~\eqref{eq:scaled_duration}--\eqref{eq:scaled_boundary_conditions_1}; they require no decision variables and can be safely paired with the empty set $\X_\sigma := \X_\tau := \emptyset$.

To each of the vertices $i \in \I$, we assign two B\'{e}zier curves: $r_i : [0,1] \rightarrow \Q_i$ (depicted in Figure~\ref{fig:bezier}) and $h_i : [0,1] \rightarrow [0, \Tmax]$.
Both these curves have a user-defined degree $d \geq \eta + 1$, where $\eta$ is the required degree of differentiability of the overall trajectory $q$.\footnote{
The assumption that the curves $r$ and $h$ have the same degree is without loss of generality.
The \emph{degree elevation} property of B\'{e}zier curves allows us to describe a B\'{e}zier curve $\gamma$ of degree $d$ as a B\'{e}zier curve $\gamma'$ of arbitrary degree $d' \geq d$, with control points that are linear functions of the control points of $\gamma$.
Any convex cost or constraint on the control points of $r$ and $h$, that takes advantage of the equal degree of these curves, can then be mapped to an equivalent convex cost or constraint on the control points of curves $r$ and $h$ of different degree.
}
The convex set $\X_i$ contains the control points of the two curves, i.e. $x_i := (r_{i,0}, \ldots, r_{i,d}, h_{i,0},\ldots, h_{i,d})$, and is defined by the following conditions:
\begin{subequations}
\label{eq:Xi}
\begin{align}
\label{eq:Xi_containment}
& r_{i,k} \in \Q_i, && k = 0, \ldots, d, \\
\label{eq:Xi_scaling}
& \dot h_{i,k} \geq \dot h_\mathrm{min}, && k = 0, \ldots, d-1, \\
\label{eq:Xi_velocity}
& \dot r_{i,k} \in \dot h_{i,k} \D, && k = 0, \ldots, d-1, \\
\label{eq:Xi_bounded}
& h_{i,0} \geq 0, \ h_{i,d} \leq T_\mathrm{max},
\end{align}
\end{subequations}
The convex constraint~\eqref{eq:Xi_containment} requires all the control points of $r_i$ to lie in the collision-free set $\Q_i$.
By the convex-hull property of the B\'{e}zier curves from Section~\ref{sec:background_on_bezier_curves}, this implies that the whole trajectory segment $r_i$ is contained in $\Q_i$.
Again by the properties of B\'{e}zier curves, the derivative $\dot h_i$ of the time scaling $h_i$ is itself a B\'{e}zier curve.
Condition~\eqref{eq:Xi_scaling} lower bounds each control point of this derivative with a small positive constant $\dot h_\mathrm{min}$ which, unless differently specified, is set to $10^{-6}$.
By the convex-hull property, this implies that $\dot h_i(s)$ is positive for all $s \in [0,1]$, and hence that $h_i$ is strictly increasing.
Since the control points of $\dot h_i$ are linear functions of the ones of $h_i$, constraint~\eqref{eq:Xi_scaling} is linear in $x_i$.
Using the definition of convexity and the positivity of $\dot h_{i,k}$, condition~\eqref{eq:Xi_velocity} can be verified to be convex in $\dot r_{i,k}$ and $\dot h_{i,k}$: this ensures that $\dot r_i (s) \in \dot h_i (s) \D$ for all $s \in [0,1]$, since the $(n+1)$-dimensional B\'{e}zier curve $(\dot r_i, \dot h_i)$ is a convex combination of the control points $(\dot r_{i,k}, \dot h_{i,k})$.
Finally, the constraints in~\eqref{eq:Xi_bounded} are conservative bounds that ensure the boundedness of $\X_i$, as assumed in Section~\ref{sec:finding_shortest_paths_in_gcs}.

We remark that asking the control points of a B\'{e}zier curve to be in a convex set is only a sufficient condition for the containment of the whole curve.
Nonetheless, the conservativeness of the conditions in~\eqref{eq:Xi} can be attenuated by increasing the degree of the curves $r_i$ and $h_i$.

\subsection{The Convex Sets $\X_e$}
\label{sec:the_convex_sets_Xe}
The first role of the edge constraints is to impose the boundary conditions~\eqref{eq:scaled_duration}--\eqref{eq:scaled_boundary_conditions_1}.
To this end, for all edges $e := (\sigma,i) \in \E$, we define $\X_e$ through the conditions $r_{i,0} = q_0$, $\dot r_{i,0} = \dot h_{i,0} \dot q_0$, and $h_{i,0} = 0$.
Given the endpoint property of B\'{e}zier curves, these linear constraints on the vector $x_i$ imply $r_i(0) = q_0$, $\dot r_i(0) = \dot h_i(0) \dot q_0$, and $h_i(0) = 0$.
Similarly, for all the edges $e:=(i,\tau)$, we define $\X_e$ via $r_{i,d} = q_T$, $\dot r_{i,d-1} = \dot h_{i,d-1} \dot q_T$, and $h_{i,d} \in [\Tmin, \Tmax]$.
For these edges, we then have $r_i(1) = q_T$, $\dot r_i(1) = \dot h_i(1) \dot q_T$, and $h_i(1) \in [\Tmin, \Tmax]$.

The second role of the edge constraints is to enforce the differentiability of the overall curves $r$ and $h$.
For all the edges $e:=(i,j) \in \E \cap \I^2$, we then define $\X_e$ through the following linear equalities:
\begin{align}
\label{eq:differentiability}
& r_{i,d-l}^{(l)} = r_{j,0}^{(l)}
\quad \text{and} \quad
h_{i,d-l}^{(l)} = h_{j,0}^{(l)},
&& l = 0, \ldots, \eta.
\end{align}
Here $r_{i,k}^{(l)}$ denotes the $k$th control point of the $l$th derivative of $r_i$.
In particular, $r_{i,k}^{(0)} = r_{i,k}$, $r_{i,k}^{(1)} = \dot r_{i,k}$, and so on.
The same notation is used for $h_i$.

\subsection{The Edge Lengths $\ell_e$}
\label{sec:the_edge_lengths_elle}
The edge lengths $\ell_e$ must reproduce the cost in~\eqref{eq:scaled_objective} by appropriately weighting the cost of each transition in the graph $G$.
This is achieved by assigning to each edge $(\sigma,i)$ outgoing the source a length of zero, and to each edge $(i,j)$ or $(i,\tau)$ the length
\begin{align}
\label{eq:edge_cost}
a (h_i(1) - h_i(0)) + b L(r_i, 1) + c E \left(\dot r_i / \sqrt{\dot h_i}, 1\right).
\end{align}
While the first term in this sum is immediately restated as the linear cost $a (h_{i,d} - h_{i,0})$, the other two terms require more work to be expressed as convex functions of $x_i$ that are amenable to efficient numerical optimization.

One option is to approximate to arbitrary precision the last two terms in~\eqref{eq:edge_cost} using numerical integration.
Since both $L$ and $E$ can be verified to be convex in the functions $r_i$ and $h_i$, the resulting expression would be convex in $x_i$, but its numerical implementation would require a large number of second-order-cone constraints, proportional to the density of the integration grid.
Instead, we prefer to minimize the following upper bounds of the last two terms in~\eqref{eq:edge_cost}:
\begin{align}
\label{eq:edge_cost_bounds}
L(r_i, 1) \leq \sum_{k=0}^{d-1} \|r_{i,k+1} - r_{i,k} \|_2
\quad \text{and} \quad
E \left(\dot r_i / \sqrt{\dot h_i}, 1\right) \leq \sum_{k=0}^{d-1} \frac{\|r_{i,k+1} - r_{i,k} \|_2^2}{h_{i,k+1} - h_{i,k}}.
\end{align}
The first inequality overestimates the length of $r_i$ by summing the distances between its control points.
The validity of this bound can be verified by applying inequality~\eqref{eq:integral_convex} to the B\'{e}zier curve $\dot r_i$ and the convex function $\|\dot r_i\|_2$.
The second inequality does a similar operation with the energy $E$, and can be checked by applying~\eqref{eq:integral_convex} to the B\'{e}zier curve $(\dot r_i, \dot h_i)$ and the function $\|\dot r_i\|_2^2/\dot h_i$, which is convex for $\dot h_i > 0$.

\subsection{Reconstruction of a Collision-Free Trajectory}
\label{sec:reconstruction_of_a_collision_free_trajectory}
Once the SPP~\eqref{eq:spp} is solved, the optimal path $p$ determines the sequence of safe regions $\Q_i$ that the robot must traverse.
To reconstruct the trajectory $r$ and the time scaling $h$, we sequence the B\'{e}zier curves $r_i$ and $h_i$ associated with these regions, as shown in Figure~\ref{fig:continuous_bezier} for $\eta := 0$.
Precisely, if the optimal path is $p := (\sigma, i_0, \ldots, i_{S-1}, \tau)$, for $\nu = 0, \ldots, S-1$, we define
\begin{align*}
&r(s) := r_{i_\nu} (s - \nu)
\quad \text{and} \quad
h(s) := h_{i_\nu} (s - \nu),
&&
\forall s \in [\nu, \nu+1].
\end{align*}

Let us verify that the functions $r$ and $h$ just defined form an optimal solution of problem~\eqref{eq:scaled}, up to the conservativeness of the constraints~\eqref{eq:Xi}  and the cost bounds~\eqref{eq:edge_cost_bounds}.
The constraints in~\eqref{eq:differentiability} imply $r, h \in \C^\eta$.
This, in turn, gives $h^{-1} \in \C^\eta$ and~\eqref{eq:scaled_continuity}.
That the collision-avoidance constraint~\eqref{eq:scaled_disjuctive} is met, is implied by~\eqref{eq:Xi_containment}.
The velocity constraint in~\eqref{eq:scaled_velocity} is verified thanks to~\eqref{eq:Xi_velocity}, while~\eqref{eq:Xi_scaling} ensures that the function $h$ is monotonically increasing, as required by the second condition in~\eqref{eq:scaled_velocity}.
The boundary conditions~\eqref{eq:scaled_duration}--\eqref{eq:scaled_boundary_conditions_1} are verified because of the constraints on the edges $(\sigma, i)$ and $(i,\tau)$ described in Section~\ref{sec:the_convex_sets_Xe}.
Finally, summing the edge lengths~\eqref{eq:edge_cost} for all the edges traversed by the path $p$ we get back~\eqref{eq:scaled_objective}.

\subsection{Class of Optimization Problems}
\label{sec:class_of_optimization_problems}
Let us conclude this section by highlighting that, by feeding the SPP in GCS we just constructed to the machinery from~\cite{marcucci2021shortest}, we obtain very tractable optimization problems.
The framework in~\cite{marcucci2021shortest} relies on \emph{perspective functions}~\cite[Section~IV.2.2]{hiriart2013convex} to handle the interplay between the discrete and continuous components of problem~\eqref{eq:spp}.
These are used to effectively ``turn off'' the edge costs $\ell_e$ and the convex constraints $(x_u, x_v) \in \X_e$ and $x_v \in \X_v$ corresponding to the edges $e$ and the vertices $v$ that do not lie along the path $p$, as required in problem~\eqref{eq:spp}.
In case of polytopic safe sets $\Q_i$ and a purely-minimum-time objective ($a:=1$ and $b:=c:=0$), it can be verified that the perspective operations lead to a mixed-integer Linear Program (LP), which, as discussed in Section~\ref{sec:rounding_the_convex_relaxation_of_the_spp}, we tackle as a simple LP followed by a rounding stage.
More generally, when the safe sets $\Q_i$ are quadratics or the cost weights $b$ and $c$ are nonzero, we obtain a mixed-integer SOCP, which we solve as a single SOCP plus rounding.
In both cases, we then have simple convex optimizations for which very efficient solvers are available (e.g. MOSEK and Gurobi).
Conversely, in order to design trajectories that are differentiable more than three times, existing MICP planners formulate prohibitive mixed-integer semidefinite programs that cannot be tackled with common solvers~\cite{deits2015efficient}.

\section{Penalties on the Higher-Order Derivatives of the Trajectory}
\label{sec:additional_differential_costs_and_constraints}
In many practical applications, we find the need to expand our problem formulation~\eqref{eq:statement} to include convex penalties on the second and higher time derivatives of the trajectory $q$.
These can be used, for example, to indirectly limit the control efforts: for a robot manipulator, in fact, the joint torques needed to execute a trajectory $q$ are proportional to the acceleration $\ddot q$ via the inertia matrix; while for a quadrotor the differential-flatness property makes the thrusts a function of the snap $q^{(4)}$~\cite{mellinger2011minimum}.
Unfortunately, even though convex in $q$, these costs become nonconvex when in problem~\eqref{eq:scaled} we optimize jointly over the shape $r$ and the time scaling $h$ of our trajectories.
While we are currently working on the design of tight convex approximations of these costs, in this section we show how simple regularization terms can be added to our SPP in GCS to prevent the higher-order derivatives of $q$ from growing excessively in magnitude.

For simplicity, let us first consider the regularization of the second derivative.
Using the chain rule, we express the acceleration $\ddot q$ in terms of the derivatives of $r$ and $h$:
$$
\ddot q(t) = \frac{\ddot r(s) - \dot q(t) \ddot h(s)}{\dot h(s)^2},
$$
where $\dot q(t) = \dot r(s)/\dot h(s)$ and $s = h^{-1}(t)$.
Using this expression, we see that a convex function of $\ddot q$ does not, in general, translate into a convex function of $r$ and $h$, and hence it cannot be directly minimized in our programs.
However, provided that we choose a bounded set $\D$ to constrain the velocity $\dot q$, the magnitude of $\ddot q$ can be kept under control by increasing the minimum value $\dot h_\mathrm{min}$ of $\dot h(s)$ and by penalizing the magnitudes of $\ddot r$ and $\ddot h$.
Letting $\varepsilon$ be a small positive scalar, a simple way to achieve the latter is the cost term
\begin{align}
\label{eq:regularizer}
\varepsilon (E(\ddot r, S) + E(\ddot h, S)),
\end{align}
which can be enforced using the ideas from Section~\ref{sec:the_edge_lengths_elle}.

The regularization of the higher-order derivatives of $q$ follows the same logic.
Specifically, using Fa\`{a} di Bruno's formula for differentiating composite functions, we see that the magnitude of $q^{(m)}$ can be regularized by increasing $\dot h_\mathrm{min}$ and by penalizing the magnitudes of $r^{(l)}$ and $h^{(l)}$, for $l=2, \ldots, m$.
The numerical results in the next section show that, even though these regularization terms are not as tight as the velocity bounds in~\eqref{eq:edge_cost_bounds}, they can sensibly smooth the trajectories we design, while only minimally affecting their cost.


\section{Numerical Results}
\label{sec:numerical_results}
We demonstrate the effectiveness of GCS on a variety of numerical examples.
In Section~\ref{sec:two_dimensional_example} we analyze a simple two-dimensional problem, and we illustrate how the different components of problem~\eqref{eq:statement} affect the shape of the trajectories we design.
In Section~\ref{sec:motion_planning_in_a_maze} we increase the environment complexity and we apply our algorithm to design paths across an intricate maze.
In Section~\ref{sec:statistical_analysis_quadrotor_flying_through_buildings}, we run a statistical analysis of the performance of GCS on the task of planning the flight of a quadrotor through randomly-generated buildings.
In Section~\ref{sec:comparison_with_prm}, we show that, with respect to widely-used sampling-based algorithms, our algorithm is capable of designing higher-quality trajectories in less runtime.
Finally, in Section~\ref{sec:coordinated_planning_of_two_robot_arms}, we demonstrate the scalability of GCS with a bimanual manipulation problem in a fourteen-dimensional configuration space.

The code necessary to reproduce all the results presented in this section can be found at~\url{https://github.com/mpetersen94/gcs}.
It uses an implementation of the SPP in GCS provided by Drake~\cite{tedrake2019drake}.
In addition to the techniques presented in~\cite{marcucci2021shortest}, the convex optimizations we solve in this paper feature additional tightening constraints, tailored to the structure of the graphs in our planning problems, and a pre-processing step that eliminates the redundancies in our graphs.
These are described in detail in Appendix~\ref{sec:further_details_on_the_implementation_of_gcs}.
The optimization solver used for the numerical experiments is MOSEK 9.2.
All experiments are run on a desktop computer with an Intel Core i7-6950X processor and 64 GB of memory.

\subsection{Two-Dimensional Example}
\label{sec:two_dimensional_example}
\begin{figure}[t]
\centering
\begin{subfigure}[t]{0.32\textwidth}
\centering
\includegraphics[width=\textwidth]{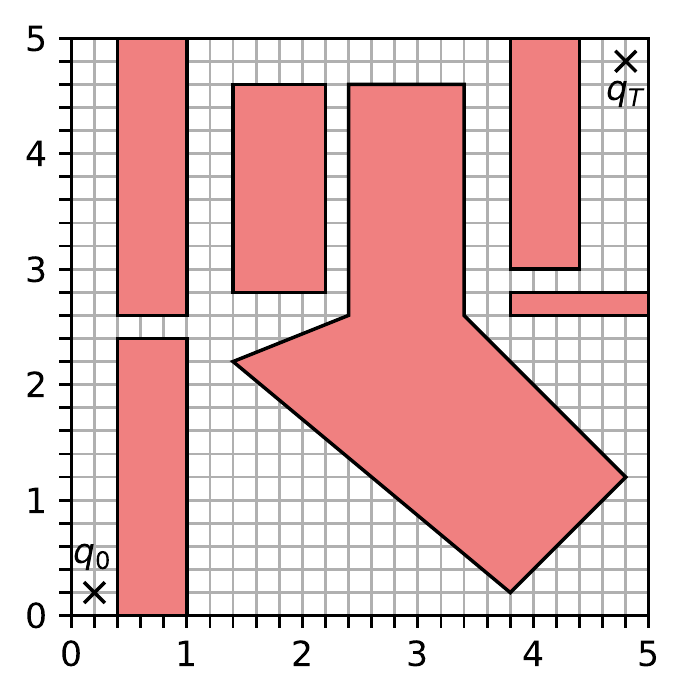}
\caption{}
\label{fig:2d_setup}
\end{subfigure}
\
\begin{subfigure}[t]{0.32\textwidth}
\centering
\includegraphics[width=\textwidth]{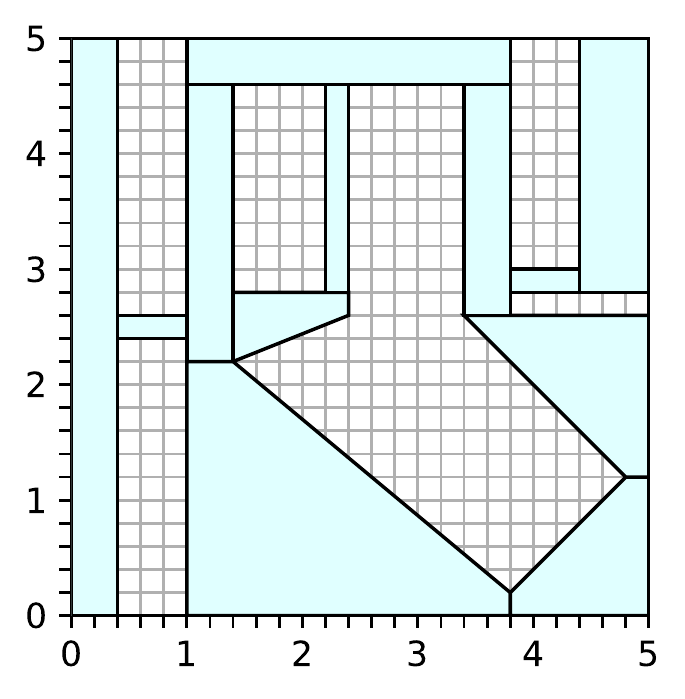}
\caption{}
\label{fig:2d_decomposition}
\end{subfigure}
\caption{Two-dimensional trajectory-design problem from Section~\ref{sec:two_dimensional_example}.
(a)~Environment with obstacles in red;  the initial $q_0$ and final $q_T$ configurations are marked with crosses.
(b)~Free space decomposed in convex safe regions $\Q_i$ (in light blue).
}
\end{figure}

\begin{figure}[t]
\centering
\begin{subfigure}[t]{0.32\textwidth}
\centering
\includegraphics[width=\textwidth]{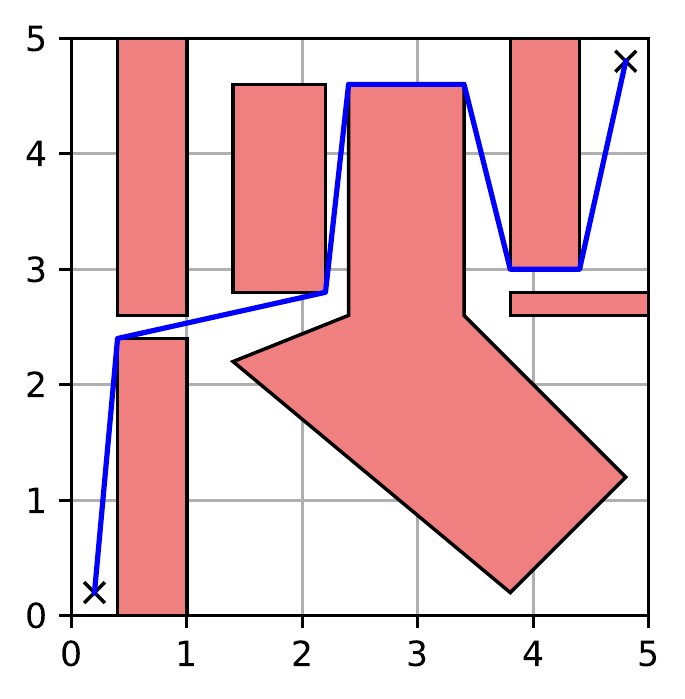}
\caption{}
\label{fig:2d_linear}
\end{subfigure}
\
\begin{subfigure}[t]{0.32\textwidth}
\centering
\includegraphics[width=\textwidth]{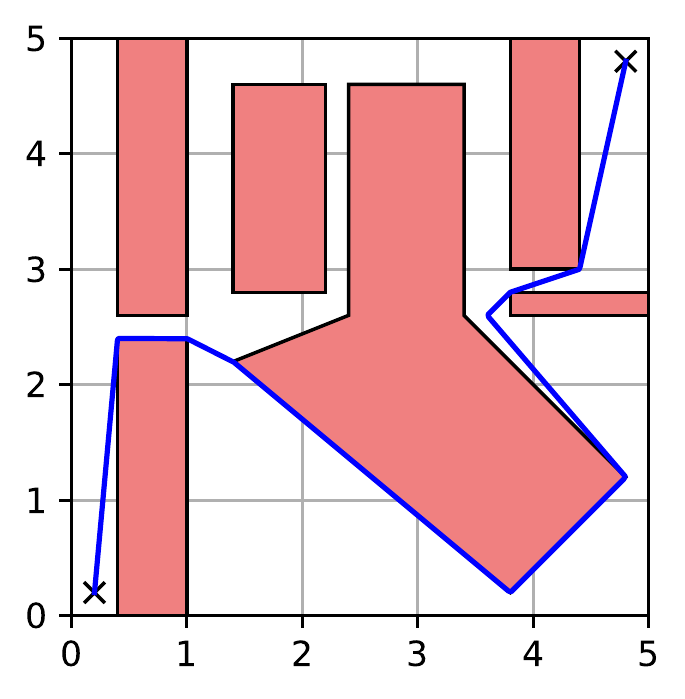}
\caption{}
\label{fig:2d_bezier_10}
\end{subfigure}
\
\begin{subfigure}[t]{0.32\textwidth}
\centering
\includegraphics[width=\textwidth]{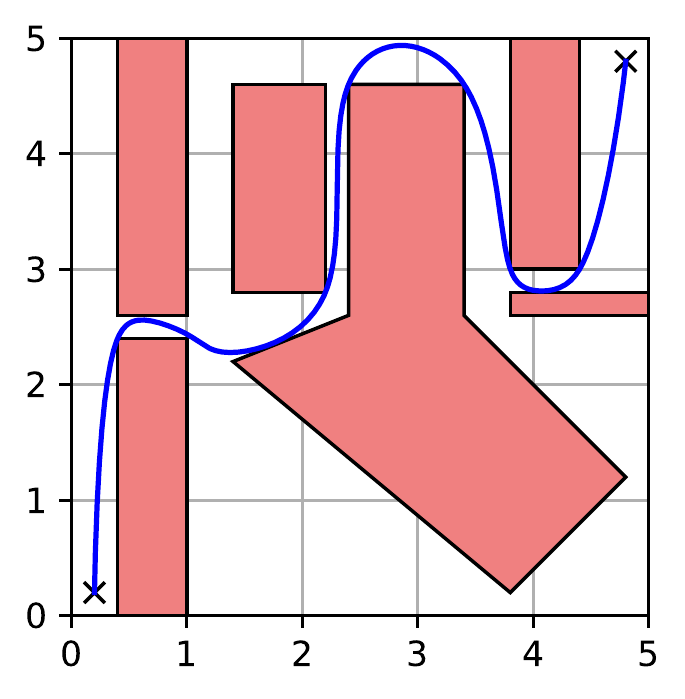}
\caption{}
\label{fig:2d_bezier_62}
\end{subfigure}
\caption{Trajectories (blue) designed by GCS for the planning problems in Section~\ref{sec:two_dimensional_example}.
(a)~Minimum-length objective.
(b)~Minimum-time objective with velocity limits $\dot q \in [-1,1]^2$.
(c)~Minimum-time objective with velocity limits, differentiability constraint $q \in \C^2$, and regularized acceleration.
GCS finds the globally-optimal trajectory ($\deltaopt = 0\%$) for each of these tasks by rounding the solution of a single convex program.
With no additional computation, it also certifies the following optimality gaps $\deltarelax$ for the rounded solutions: $1.7\%$ for (a), $7.3\%$ for (b), and $3.0\%$ for (c).
}
\end{figure}

\begin{figure}[h!]
\centering
\begin{subfigure}[t]{0.42\textwidth}
\centering
\includegraphics[width=\textwidth]{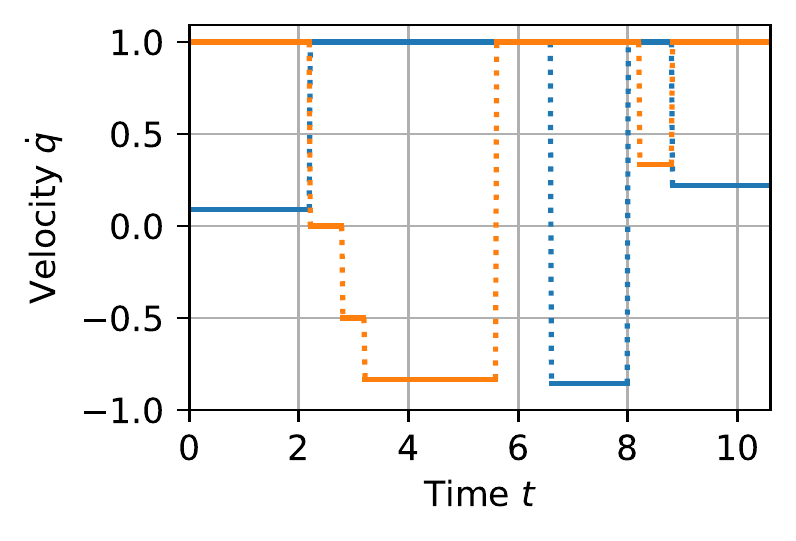}
\caption{}
\label{fig:2d_bezier_10_vel}
\end{subfigure}
\
\begin{subfigure}[t]{0.42\textwidth}
\centering
\includegraphics[width=\textwidth]{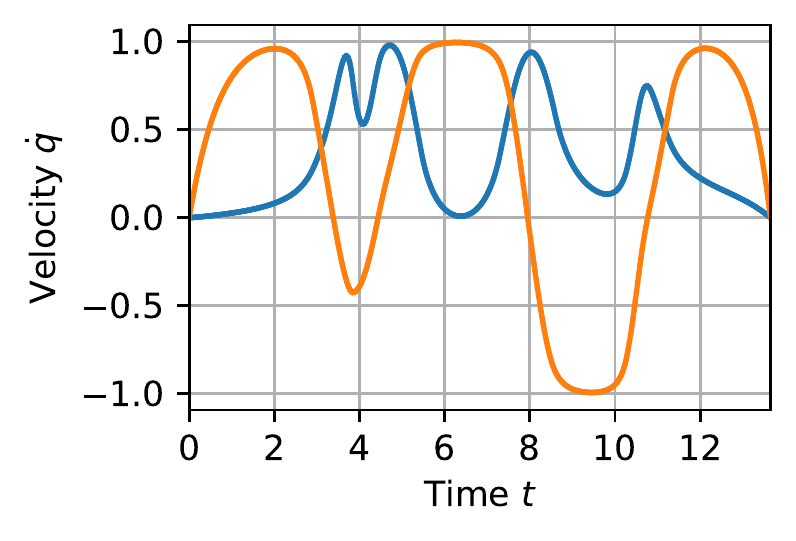}
\caption{}
\label{fig:2d_bezier_62_vel}
\end{subfigure}
\caption{(a)~Velocity profile for the minimum-time trajectory depicted in Figure~\ref{fig:2d_bezier_10}, with dotted lines representing discontinuities.
(b)~Velocity profile for the smoothed trajectory in Figure~\ref{fig:2d_bezier_62}.
The horizontal component of $\dot q$ is blue, the vertical is orange.
In both the problems, the velocity components are constrained to lie in the interval $[-1,1]$.
}
\end{figure}

The goal of our first numerical example is to illustrate how the different parameters in problem~\eqref{eq:statement} affect the shape of the trajectories we design.
To this end, we consider the simple two-dimensional environment depicted in Figure~\ref{fig:2d_setup}.
The initial $q_0 := (0.2,0.2)$ and final $q_T := (4.8,4.8)$ configurations are marked with a black cross; the obstacles are the red polygons.
The convex decomposition $\{\Q_i\}_{i \in \I}$ of the free space $\Q$ is depicted in light blue in Figure~\ref{fig:2d_decomposition}.

The first planning problem we analyze asks to minimize the total Euclidean length of the trajectory.
The weights in the objective~\eqref{eq:statement_objective} are then $a:=c:=0$ and $b:=1$.
The trajectory $q$ is only required to be continuous ($\eta:=0$), while velocity and time constraints are irrelevant for a minimum-length problem.
We let the degree of the B\'{e}zier curves $r$ and $h$ be $d:=1$ (i.e. straight lines).
Solving the convex relaxation of the SPP in GCS we obtain the cost $\Crelax = 10.77$, while the rounding step from Section~\ref{sec:rounding_the_convex_relaxation_of_the_spp} gives us the feasible trajectory depicted in Figure~\ref{fig:2d_linear} with cost $\Cround = 10.96$.
By comparing these two numbers, GCS automatically provides the optimality bound $\deltarelax := (\Cround - \Crelax)/\Crelax = 1.7\%$ for the rounded solution.
However, by actually running a slightly more expensive mixed-integer solver, it is possible to verify that the rounded solution is indeed the global minimizer: $\Cround = \Copt$ and $\deltaopt := (\Cround - \Copt)/\Copt = 0\%$.

For the second scenario, we consider a minimum-time problem with velocity limits.
The weights in problem~\eqref{eq:statement} are set to $a:=1$ and $b:=c:=0$.
We look for a continuous trajectory ($\eta:=0$), whose velocity $\dot q$ is contained in the box $\D:= [-1,1]^2$ for all times $t$.
The bounds on the trajectory duration are set to $\Tmin \approx 0$ and $\Tmax \gg 0$, so that they do not affect the optimization problem.
For this problem we let the optimizer decide the initial $\dot q(0)$ and final $\dot q(T)$ velocities by dropping the boundary conditions~\eqref{eq:statement_boundary_conditions_1}.
Once again, we use B\'{e}zier curves of degree $d:=1$.
The trajectory generated by GCS is illustrated in Figure~\ref{fig:2d_bezier_10}.
The convex relaxation has cost $\Crelax = 9.88$, while the rounded trajectory has duration $\Cround=10.60$ and, therefore, it is certified to be within $\deltarelax = 7.3\%$ of the global minimum.
As before, a mixed-integer solver can be used to verify that the trajectory generated by GCS is actually globally optimal ($\deltaopt = 0\%$).

In juxtaposition to the minimum-length case, the minimum-time trajectory in Figure~\ref{fig:2d_bezier_10} passes below the central obstacle.
This is because, although shorter, the trajectory in Figure~\ref{fig:2d_linear} is everywhere almost horizontal or vertical, and in these directions the speed is limited by the constraint set $\D$ to $\|\dot q\|_2 \leq 1$.
The route below the obstacle is slightly longer, but it allows diagonal motion with speed $\|\dot q\|_2 \leq \sqrt 2$.
In Figure~\ref{fig:2d_bezier_10_vel} we report the velocity $\dot q$ corresponding to the minimum-time trajectory: as expected, the optimal velocity is discontinuous and, at all times $t$, either the horizontal or the vertical component of $\dot q$ reaches the upper bound of $1$.

Finally, we show the effects of the regularization strategy discussed in Section~\ref{sec:additional_differential_costs_and_constraints} on the smoothness of the minimum-time trajectory.
We require the curve $q$ to be twice continuously differentiable ($\eta:=2$).
The initial and final velocities are forced to be zero, $\dot q_0 := \dot q_T := 0$, and we set the degree of the B\'{e}zier curves to $d:=6$.
We increase $\dot h_\mathrm{min}$ from its default value of $10^{-6}$ to $10^{-1}$, and we add the penalty~\eqref{eq:regularizer} with weight $\varepsilon = 10^{-1}$.
The resulting trajectory is reported in Figures~\ref{fig:2d_bezier_62} and~\ref{fig:2d_bezier_62_vel}.
As can be seen, the regularization smooths the optimal trajectory significantly and even changes its homotopy class.
The costs of the convex relaxation and the rounded solution increase to $\Crelax = 27.29$ and $\Cround = 28.10$, respectively.
The optimality gap certified by GCS is hence $\deltarelax = 3.0\%$, but, once again, a mixed-integer solver can be used to verify that the rounded solution is actually globally optimal ($\deltaopt = 0\%$).
The duration of the smoothed trajectory is $T=13.65$.

\subsection{Motion Planning in a Maze}
\label{sec:motion_planning_in_a_maze}
\begin{figure}[t]
\centering
\begin{subfigure}[t]{0.48\textwidth}
\centering
\includegraphics[width=\textwidth]{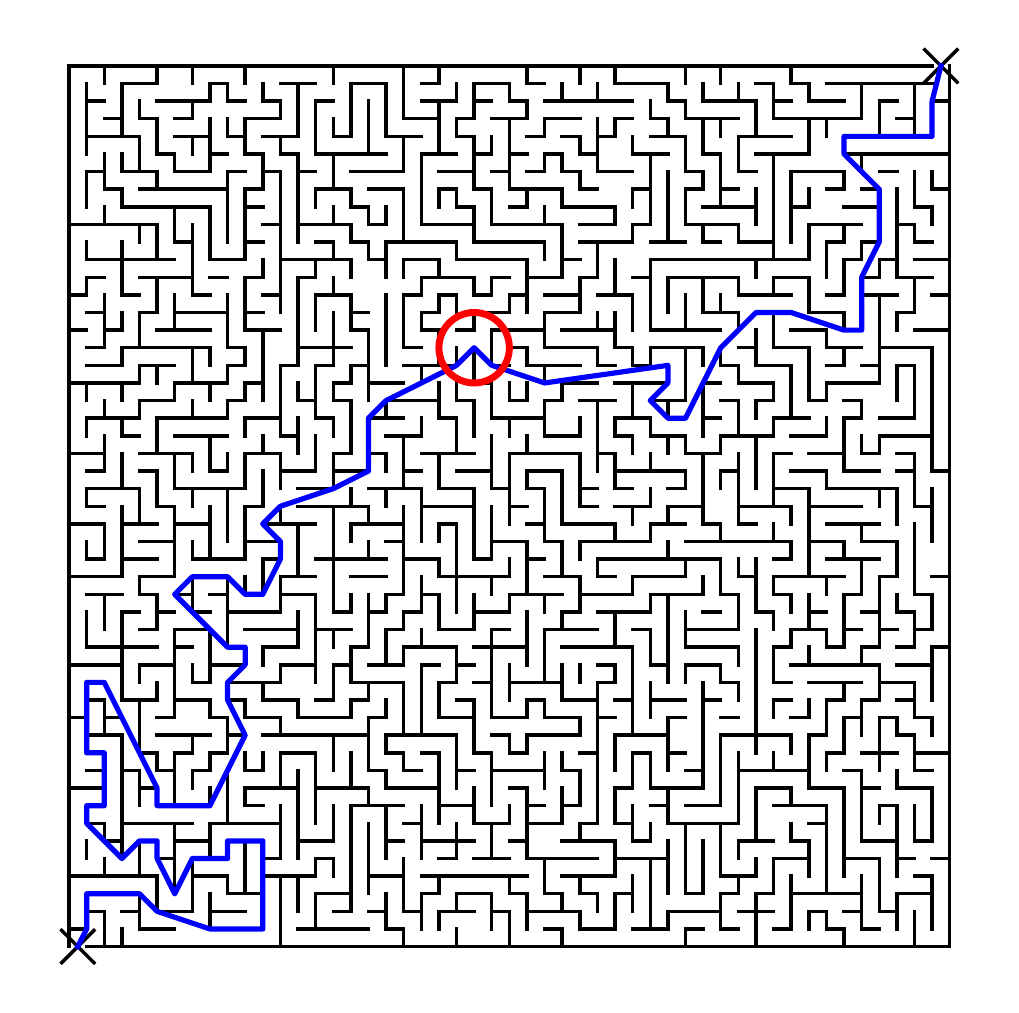}
\caption{}
\label{fig:maze_length}
\end{subfigure}
\
\begin{subfigure}[t]{0.48\textwidth}
\centering
\includegraphics[width=\textwidth]{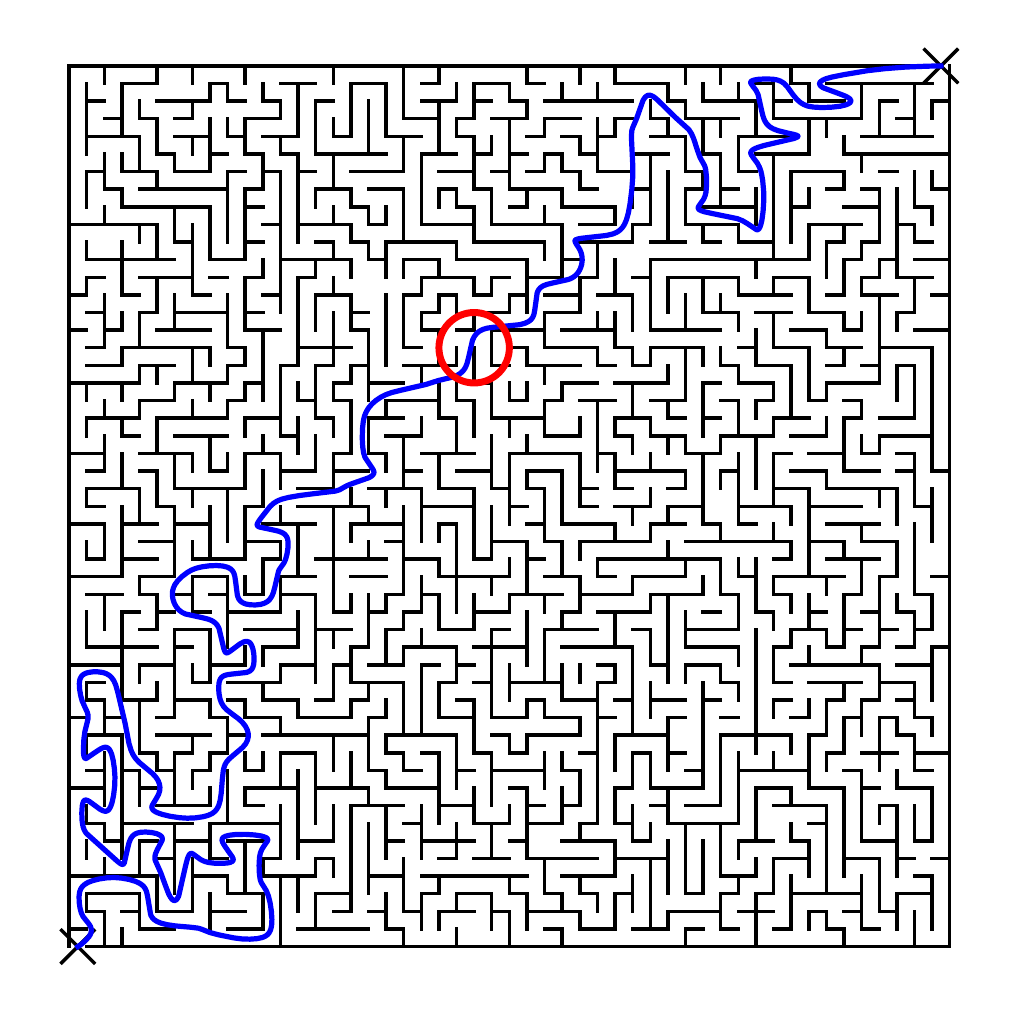}
\caption{}
\label{fig:maze_time}
\end{subfigure}
\caption{Solutions of the motion-planning problems through a maze from Section~\ref{sec:motion_planning_in_a_maze}.
(a)~Minimum-length trajectory connecting the start (bottom-left cross) and the goal (top-right cross).
(b)~Solution of the minimum-time problem with velocity constraint $\dot q \in [-1,1]^2$ and regularized acceleration.
The solutions of these two problems bifurcate at the red circle, and take different paths across the maze.
For both problems, GCS identifies the globally-optimal trajectory via a single SOCP.
}
\label{fig:maze}
\end{figure}

In this example we consider a two-dimensional planning problem of higher complexity than the one just analyzed: we design trajectories through the maze depicted in Figure~\ref{fig:maze}.
The maze has $50 \cdot 50=2,500$ cells.
The starting cell is the one at the bottom left, the goal cell is in the top right.
The graph of convex sets is constructed by making each cell into a safe set $\Q_i$.
Bidirectional edges are drawn between cells that are not separated by a wall.
The maze is generated using random depth-first search.
Since mazes constructed using this algorithm have all cells connected to the starting cell by a unique path, to make the planning problem more challenging, we create multiple paths to the goal by randomly selecting and removing $100$ walls from the maze.

As in the previous example, we consider a minimum-length problem and a minimum-time problem with regularized acceleration.
We set the parameters $(a, b, c, \eta, \D, \Tmin, \Tmax, \dot q_0, \dot q_T, d)$ to the same values we adopted in the corresponding problems in Section~\ref{sec:two_dimensional_example}.
The optimal trajectories across the maze corresponding to the two objective functions are reported in Figures~\ref{fig:maze_length} and~\ref{fig:maze_time}.
As it can be seen, the two curves visit different sequences of safe sets (cells).
In particular, the sharp turn taken by the minimum-length trajectory, circled in red in Figure~\ref{fig:maze_length}, would be expensive for the second problem, where we have a penalty on the magnitude of the acceleration.
For the curve in Figure~\ref{fig:maze_time}, GCS decides then to take a longer but smoother route to the goal.
For both the problems under analysis, the convex relaxation returns a solution with binary probabilities $\varphi_e$.
Rounding is then unnecessary in this case, and the solution of the convex relaxation is automatically certified to be globally optimal ($\deltarelax = \deltaopt = 0\%$).

We find this example powerful because it highlights the transparency with which GCS blends discrete and continuous optimization.
Finding a discrete sequence of cells to traverse the maze in Figure~\ref{fig:maze} is a trivial graph search.
Also finding a path of minimum length, as in Figure~\ref{fig:maze_length}, is a relatively simple problem: in fact, in two dimensions, the Euclidean SPP is solvable in polynomial time by constructing a discrete visibility graph~\cite{lozano1979algorithm}.
On the other hand, designing a trajectory like the minimum-time one in Figure~\ref{fig:maze_time} is a substantially more involved operation.
GCS gives us a unified mathematical framework that can tackle all these problems very efficiently, while embracing both the higher-level combinatorial structure and the lower-level convexity of our planning problems.

In conclusion of this example let us illustrate another axis in which GCS significantly improves on existing MICP planners.
The worst-case runtime of a mixed-integer solver is typically exponential in the number of binaries in the optimization problem.
Previous MICP planners parameterize a single trajectory and subdivide it in a fixed number of segments, then they use a binary variable to assign each segment to each safe region $\Q_i$~\cite{deits2015efficient}.
Given that, in the worst case, the optimal trajectory might visit all the safe regions $\Q_i$, this approach requires a total of $|\I|^2$ binary variables.
For the maze in Figure~\ref{fig:maze}, we would then have $|\I|^2 = 2,500^2 = 6.25 \cdot 10^6$ binaries: a quantity well beyond the capability of today's solvers.
On the contrary, GCS uses only two binaries per pair of intersecting regions, and it yields an MICP with only $5198 \approx 2 |\I|$ binaries, which is solved exactly through a single SOCP.

\subsection{Statistical Analysis: Quadrotor Flying through Buildings}
\label{sec:statistical_analysis_quadrotor_flying_through_buildings}
\begin{figure}[t]
\centering
\includegraphics[width=\textwidth]{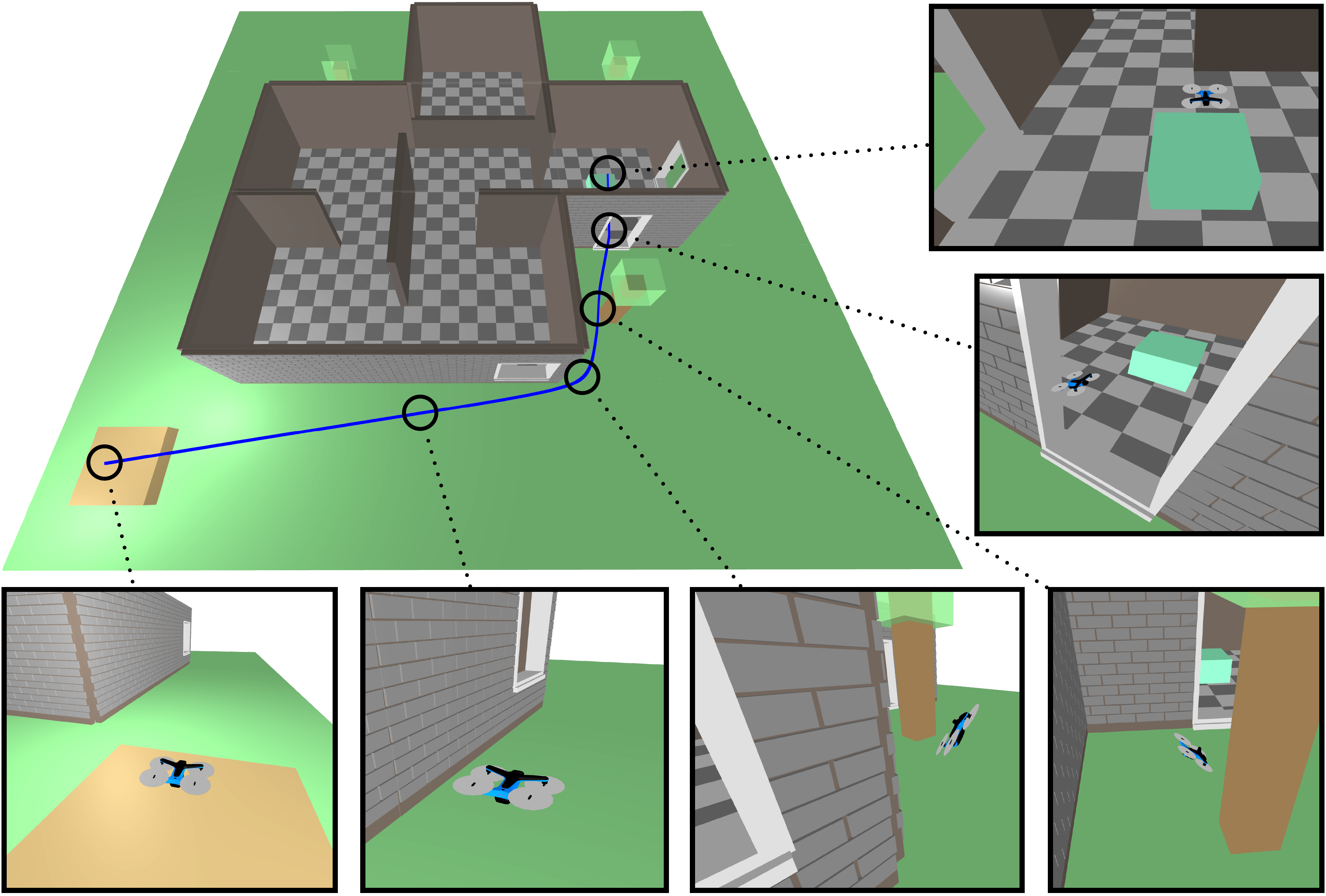}
\caption{
One of the randomly-generated environments for the statistical analysis in Section~\ref{sec:statistical_analysis_quadrotor_flying_through_buildings}.
The trajectory generated by GCS for the center of mass of the quadrotor is depicted in blue.
The robot orientation is reconstructed taking advantage of the differential flatness of the system dynamics.
The snapshots show the starting and ending configurations, as well as the quadrotor flying close to the obstacles in the environment.
}
\label{fig:quadrotor}
\end{figure}

In this section we present a statistical analysis of the performance of GCS.
Taking inspiration from~\cite{deits2015efficient}, we test our algorithm on the task of planning the motion of a quadrotor through randomly-generated buildings.
An example of such a task is illustrated in Figure~\ref{fig:quadrotor}: while moving from the brown to the green block, the quadrotor needs to fly around trees, and through doors and windows.
A brief description of how the buildings are generated can be found in Appendix~\ref{sec:random_environment_generation}.

Even though the configuration space of a quadrotor is six dimensional, the differential-flatness property of this system allows us to plan dynamically-feasible trajectories directly in the three-dimensional Cartesian space.
In fact, given a four-times-differentiable trajectory of the position of the center of mass, the time evolution of the quadrotor's orientation, together with the necessary control signals, is uniquely defined and easily computed~\cite{mellinger2011minimum}.
The space in which we design trajectories is then $\Q \subset \R^3$ and, given that all the obstacles have polyhedral shape (as in Figure~\ref{fig:quadrotor}), the decomposition of this space into convex sets $\Q_i$ can be done exactly.
Appendix~\ref{sec:random_environment_generation} provides more details on this decomposition.

In the formulation of the planning problem~\eqref{eq:statement}, we penalize with equal weight the duration and the length of the trajectory ($a:=b:=1$ and $c:=0$).
We parameterize the trajectories using B\'{e}zier curves of degree $d := 7$ and, to take advantage of the differential flatness, we require these curves to be continuously differentiable $\eta := 4$ times.
The velocity is constrained to be in the box $\D:= [-10,10]^3$ for all times.\footnote{
To contextualize the velocity limits, consider that the random environments are squares with sides of length 25, and the collision geometry of the quadrotor is a sphere of radius 0.2 (see also Appendix~\ref{sec:random_environment_generation}).
}
The limits $\Tmin$ and $\Tmax$ on the duration of the trajectory have values that do not affect the optimal solution.
As said, the initial $q_0$ and final $q_T$ positions are above the brown and green boxes, respectively.
The boundary values of the velocity are zero $\dot q_0 := \dot q_T := 0$, as well as the boundary values of the second and the third derivatives of the trajectory.\footnote{
For $l = 2, \ldots, L$, the derivative constraints $q^{(l)}(0) = q^{(l)}(T) = 0$ in problem~\eqref{eq:statement} map to the conditions $r^{(l)}(0) = h^{(l)}(0) \dot q_0$ and $r^{(l)}(S) = h^{(l)}(S) \dot q_T$ in problem~\eqref{eq:scaled}.
The latter are linear in the decision variables of our SPP in GCS, and can be easily incorporated among the edge constraints listed in Section~\ref{sec:the_convex_sets_Xe}.
}
Because of the differential flatness, the latter ensure that the quadrotor starts and ends the motion with horizontal orientation and zero angular velocity.
Finally, to regularize the acceleration of the quadrotor, as discussed in Section~\ref{sec:additional_differential_costs_and_constraints}, we set $\dot h_\mathrm{min} := 10^{-3}$.

\begin{figure}[t]
\centering
\begin{subfigure}[t]{0.48\textwidth}
\centering
\includegraphics[width=\textwidth]{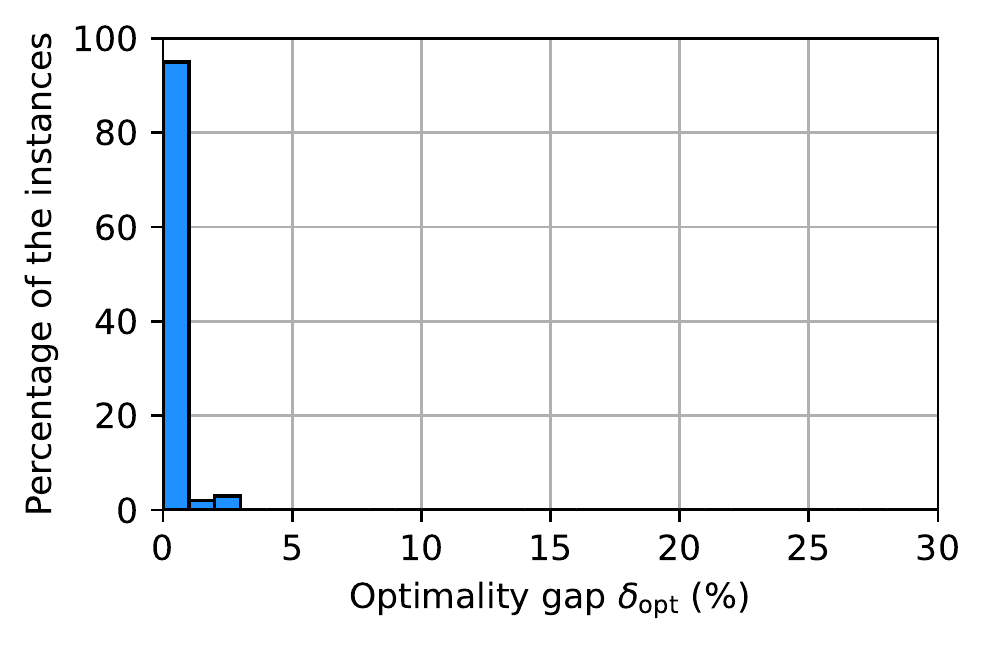}
\caption{}
\label{fig:delta}
\end{subfigure}
\hfill
\begin{subfigure}[t]{0.48\textwidth}
\centering
\includegraphics[width=\textwidth]{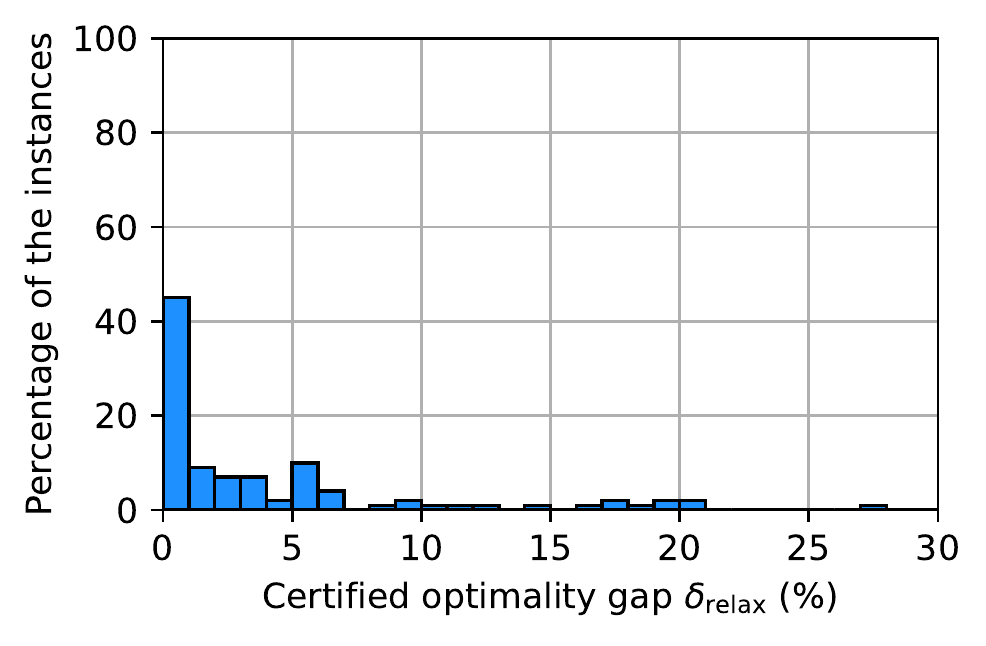}
\caption{}
\label{fig:delta_certified}
\end{subfigure}
\caption{
Histograms of the optimality gaps registered in the statistical analysis in Section~\ref{sec:statistical_analysis_quadrotor_flying_through_buildings}.
(a)~Optimality gap $\deltaopt$: percentage gap between the cost of the solution returned by GCS and the global optimum.
On $95\%$ of the environments GCS designs a trajectory with optimality gap smaller than $1\%$, and, even in the worst case, it finds a solution whose cost is only $2.9\%$ larger than the global minimum.
(b)~Optimality gap $\deltarelax \geq \deltaopt$ automatically certified by GCS.
On $68\%$ (respectively $84\%$) of the problems GCS certifies that the returned solution has optimality gap smaller than $4\%$ (respectively $7\%$).
}
\label{fig:quadrotor_histograms}
\end{figure}

We plan the motion of the quadrotor through 100 random buildings.
To assess the quality of the trajectories generated by GCS, we look at the optimality gaps $\deltaopt$ and $\deltarelax$.
As in the previous examples, the value of $\deltaopt$ is computed (just for analysis purposes) by solving the planning problem to global optimality using a mixed-integer algorithm, while $\deltarelax$ is the upper bound on $\deltaopt$ that is automatically provided to us by GCS.
The histograms of these two quantities across the 100 experiments are reported in Figure~\ref{fig:quadrotor_histograms}.
Figure~\ref{fig:delta} shows that on $95\%$ of the environments GCS designs a trajectory whose optimality gap $\deltaopt$ is smaller than $1\%$, and, even in the worst case, is only $2.9\%$.
From Figure~\ref{fig:delta_certified}, we see that on $68\%$ (respectively $84\%$) of the problems GCS certifies that the returned solution is within $4\%$ (respectively $7\%$) of the global optimum.
The largest optimality gap $\deltarelax$ certified by GCS is $27.1\%$, and it corresponds to an environment where we have $\deltaopt = 2.3\%$.
Therefore, even for this problem instance, the moderately-large value of $\deltarelax$ is mostly due to the convex relaxation being slightly loose, rather than the rounded solution being suboptimal.

We report that, for the statistical analysis in this subsection, we set the MOSEK parameter \texttt{MSK\_IPAR\_INTPNT\_SOLVE\_FORM = 1}, which tells the interior-point solver to interpret our optimizations in standard primal form~\cite{lofberg2009dualize}.
Without this, MOSEK encountered numerical issues in the solution of the convex relaxations of the motion-planning problems.
This parameter choice has the drawback of sensibly slowing down the planning times: the solve times for the convex relaxations of the 100 motion plans have median $3.7$~s, mean $6.4$~s, and maximum $31.2$~s.
However, we are very confident that a deeper analysis of these numerical issues and a tailored pre-solve stage, can reduce these times by at least one order of magnitude.


\subsection{Comparison with PRM: Motion Planning of a Robot Arm}
\label{sec:comparison_with_prm}
In this subsection we consider the motion planning of a robot arm, and we compare GCS with commonly-used sampling-based planners.
GCS is a multiple-query algorithm, meaning that the same data structure (the graph of convex sets) can be used to plan the motion of the robot for many initial and final conditions.
Its natural sampling-based comparison is then the Probabilistic-RoadMap (PRM) algorithm~\cite{kavraki1996probabilistic}.
The robot arm we use in this benchmark is the KUKA LBR iiwa with $n=7$ degrees of freedom: we have chosen a seven-dimensional configuration space $\Q$ since PRM methods can struggle in larger spaces, and both algorithms under analysis can easily design trajectories in lower dimensions.

\begin{figure}[t]
\centering
\begin{subfigure}[b]{.3\textwidth}
\centering
\includegraphics[width=\textwidth]{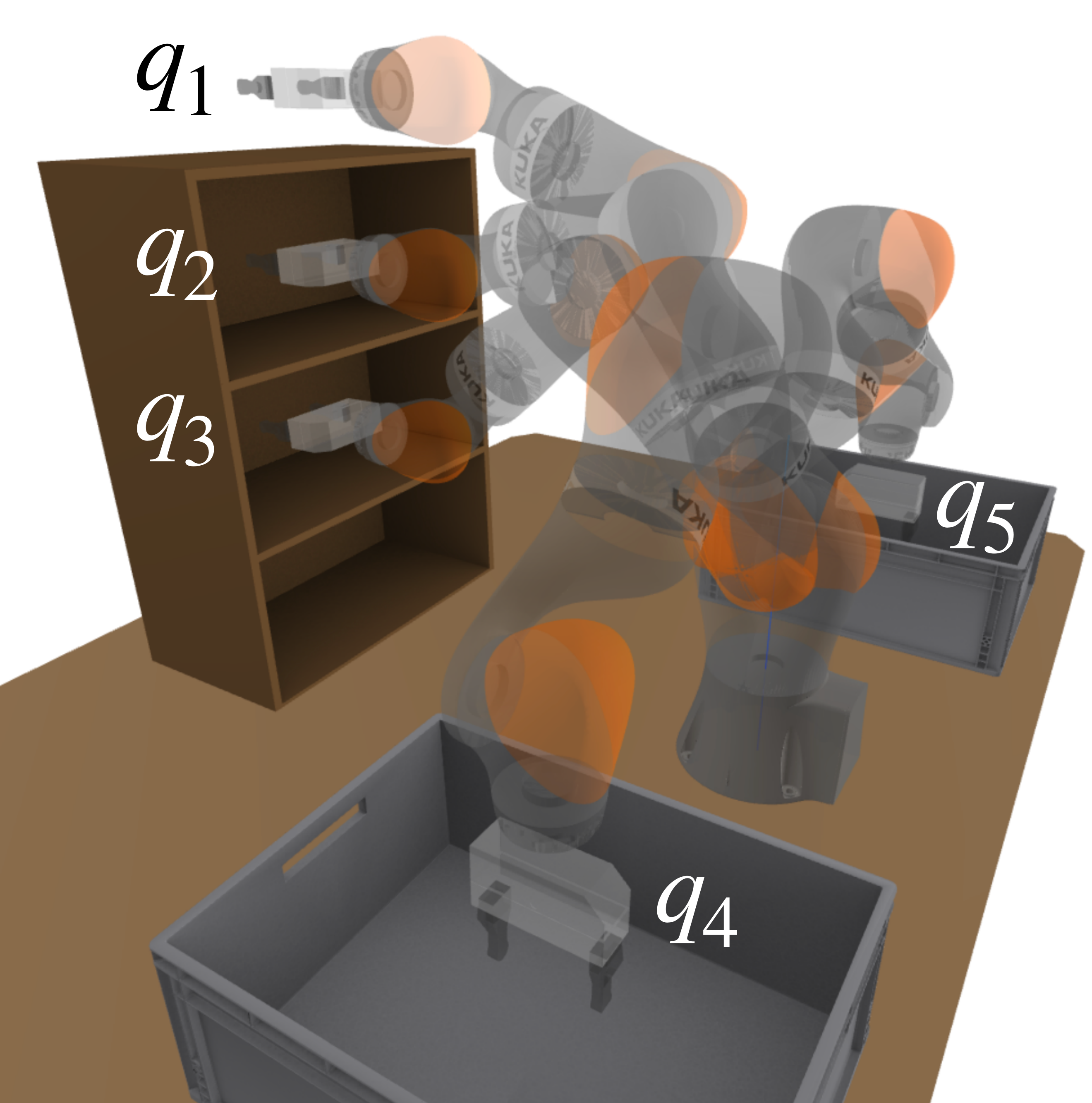}
\caption{}
\label{fig:narrow_passages_seeds}
\end{subfigure}
\begin{subfigure}[b]{.3\textwidth}
\centering
\includegraphics[width=\textwidth]{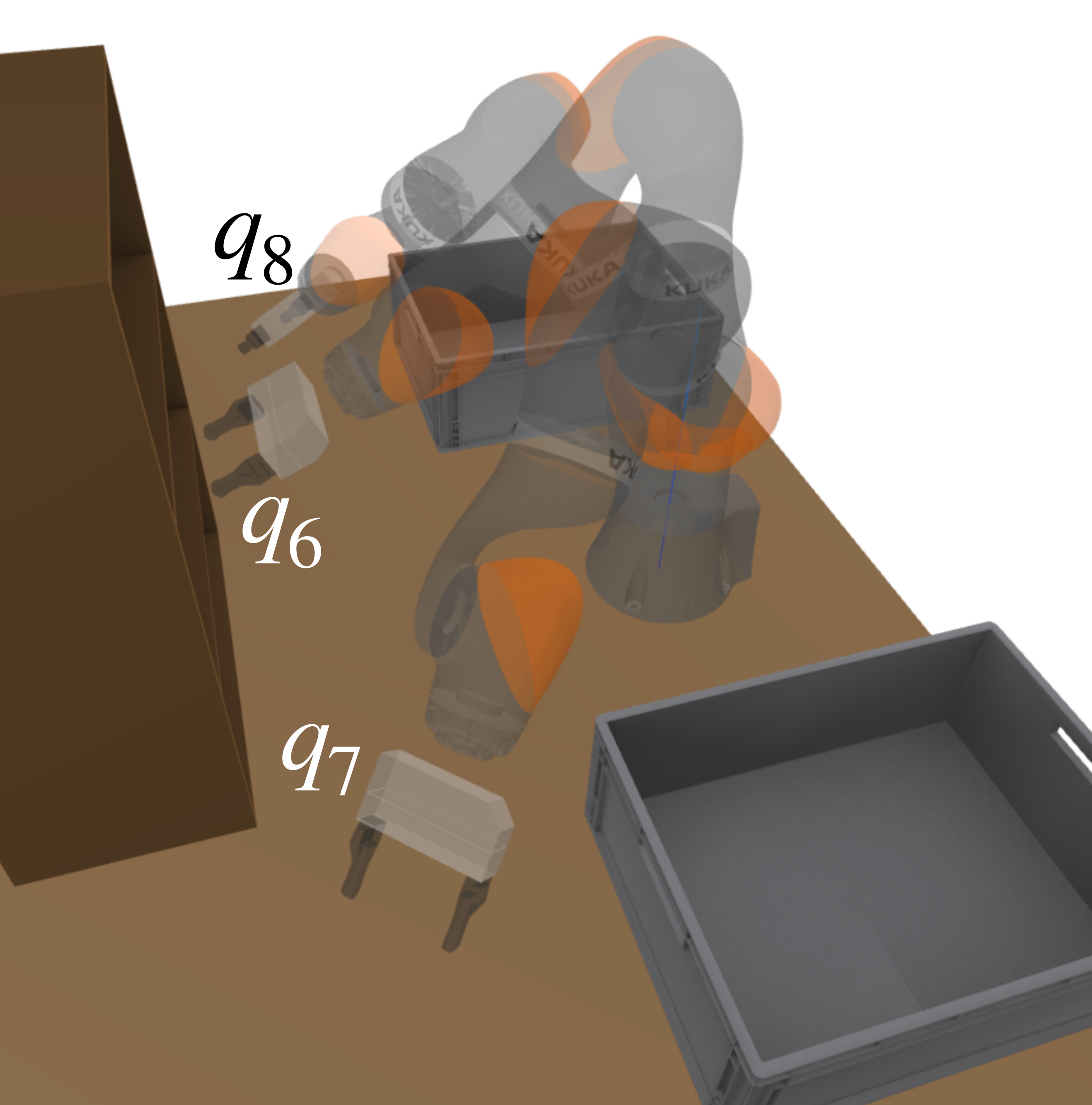}
\caption{}
\label{fig:free_space_seeds}
\end{subfigure}
\begin{subfigure}[b]{.32\textwidth}
\centering
\includegraphics[width=\textwidth]{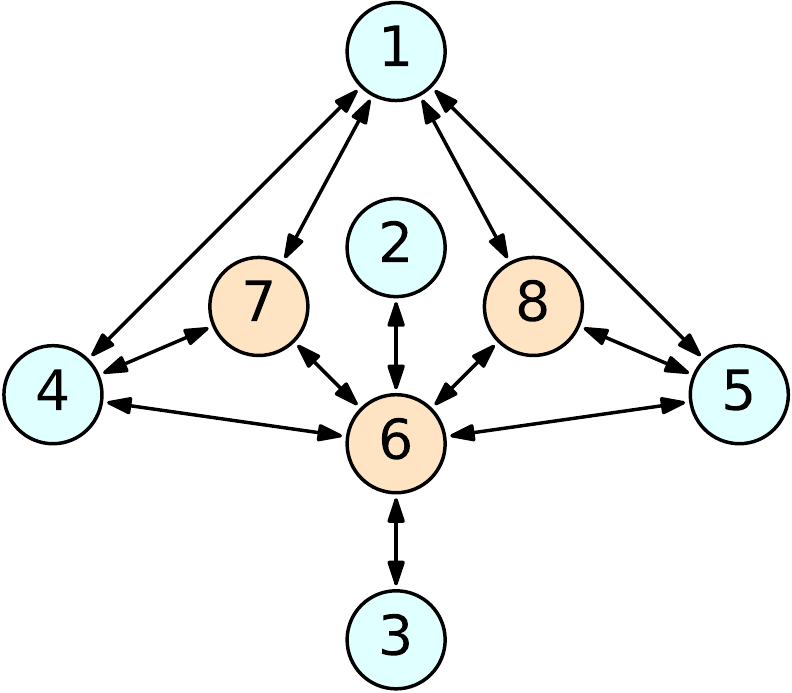}
\caption{}
\label{fig:graph}
\end{subfigure}
\caption{
Construction of the GCS for the motion planning of the robot arm in Section~\ref{sec:comparison_with_prm}.
(a)~Five seed poses $\{q_k\}_{k=1}^5$ for the region-inflating algorithm from~\cite{deits2015computing}.
These are chosen to fill the space within the rack and the bins.
(b)~The remaining three seed poses $\{q_k\}_{k=6}^8$, chosen to approximately fill the free configuration space.
(c)~The graph $G$ obtained by processing the safe regions $\Q_i$.
The light-blue vertices correspond to the seed poses in (a), the light-brown ones to the additional seeds from (b).
Vertices are labeled with the subscripts of the corresponding poses.
}
\label{fig:kuka_graph}
\end{figure}

The robot arm is depicted in Figure~\ref{fig:kuka_graph}, and it is required to move within an environment composed of a rack (in front of the robot) and two bins (on the sides). 
As opposed to the examples considered so far, an exact decomposition of the free configuration space $\Q$ is not feasible in this application.
We then adopt the approximate decomposition algorithm, IRIS, from~\cite{deits2015computing}; more precisely, its extension to configuration spaces with nonconvex obstacles, \href{https://drake.mit.edu/doxygen_cxx/group__geometry__optimization.html#ga3a51e0fec449a0abcf498f78a2a390a8}{\texttt{IrisInConfigurationSpace}}, implemented in Drake~\cite{tedrake2019drake}.
Given a ``seed pose'' of the robot, this algorithm inflates a polytope of robot configurations that are not in collision with the environment.
While these polytopes could be rigorously certified to be collision free~\cite{amice2022finding}, for the experiments reported here we use a fast implementation based on nonconvex optimization that does not provide a rigorous certification, but that appears to be very reliable in practice.

Automatic seeding of the regions is certainly possible, but we have found that producing seeds manually via inverse kinematics, together with a simple visualization of the graph $G$ to check the connectivity between regions $\Q_i$, is straightforward and highly effective.
We use IRIS to construct a total of eight safe polytopes $\Q_i$, whose corresponding seed poses $q_i$ are depicted in Figures~\ref{fig:kuka_graph}.
The seed poses $\{q_i\}_{i=1}^5$ in Figures~\ref{fig:narrow_passages_seeds} are chosen to create polytopes $\Q_i$ that cover the volume of configuration space for which the end effector is in the vicinity of the rack and the bins.
The poses $\{q_i\}_{i=6}^8$ in Figures~\ref{fig:free_space_seeds} are picked to approximately fill the rest of the free space.
The construction of the safe regions is parallelized, and took us 53 seconds.
By processing the safe regions $\Q_i$ as described in Section~\ref{sec:the_graph_G}, we obtain the graph $G$ depicted in Figure~\ref{fig:graph}.
The vertices $\I = \{1, \ldots, 8\}$ are the subscripts of the poses that we use as seeds for the construction of each polytope, i.e., vertex $i \in \I$ is paired with the safe polytope $\Q_i$ obtained from the seed $q_i$.
As can be seen from the connectivity of the graph, the polytopes $\Q_i$ are sufficiently inflated to connect all the seed poses $q_i$.
At runtime, given the initial $q_0$ and final $q_T$ configuration, the source $\sigma$ and the target $\tau$ vertices are added to the graph and connected to other vertices as described in Section~\ref{sec:the_graph_G}.

In practice, the plans generated by a PRM can be very suboptimal and are rarely commanded to the robot directly.
While asymptotically-optimal versions of the PRM method exist~\cite{karaman2011sampling}, in our experience, in the relatively high-dimensional space we consider here, the increase in performance of these variants is not worth their computational cost.
A solution commonly used in practice is then to post-process the plans generated by the PRM with a simple short-cutting algorithm.
This algorithm samples pairs of points along the PRM trajectory and connects them via straight segments: if a segment is verified to be collision free the trajectory is successfully shortened.
This step can dramatically shorten the PRM trajectories but it requires time-consuming collision checks: for this reason, here we compare GCS with both the regular PRM and the PRM with short-cutting.
For both the PRM methods we use the implementation from~\cite{cru}.
More implementation details can be found in Appendix~\ref{sec:implementation_of_the_prm}; here we only mention that our roadmap is composed of $15 \cdot 10^3$ sample configurations and its construction took, with our (not fully optimized) setup, 16~minutes.

\begin{figure}[t]
\centering
\begin{subfigure}[b]{.32\textwidth}
\includegraphics[width=\textwidth]{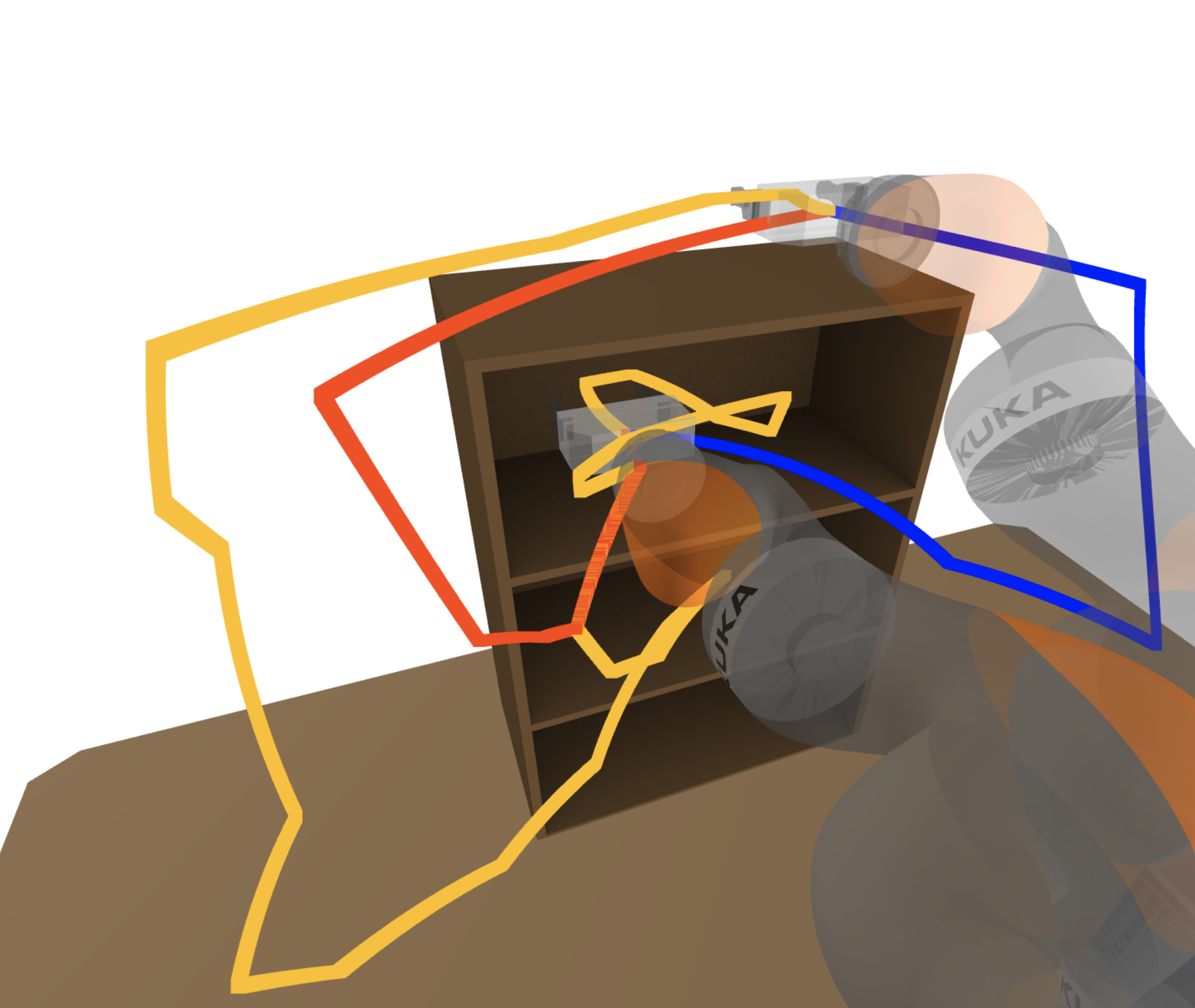}
\caption{}
\label{fig:plan_1}
\end{subfigure}
\
\begin{subfigure}[b]{.32\textwidth}
\includegraphics[width=\textwidth]{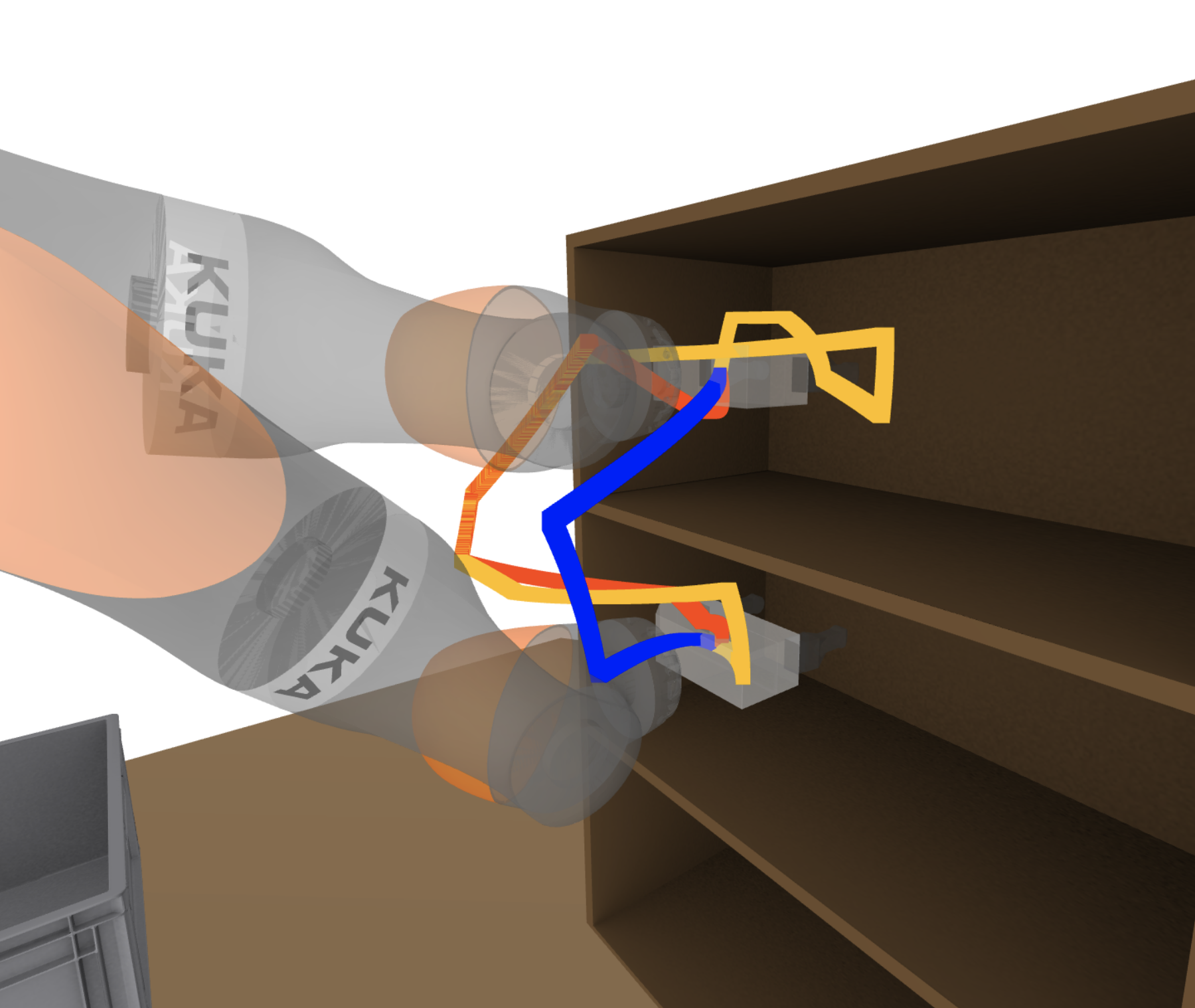}
\caption{}
\label{fig:plan_2}
\end{subfigure}
\
\begin{subfigure}[b]{.32\textwidth}
\includegraphics[width=\textwidth]{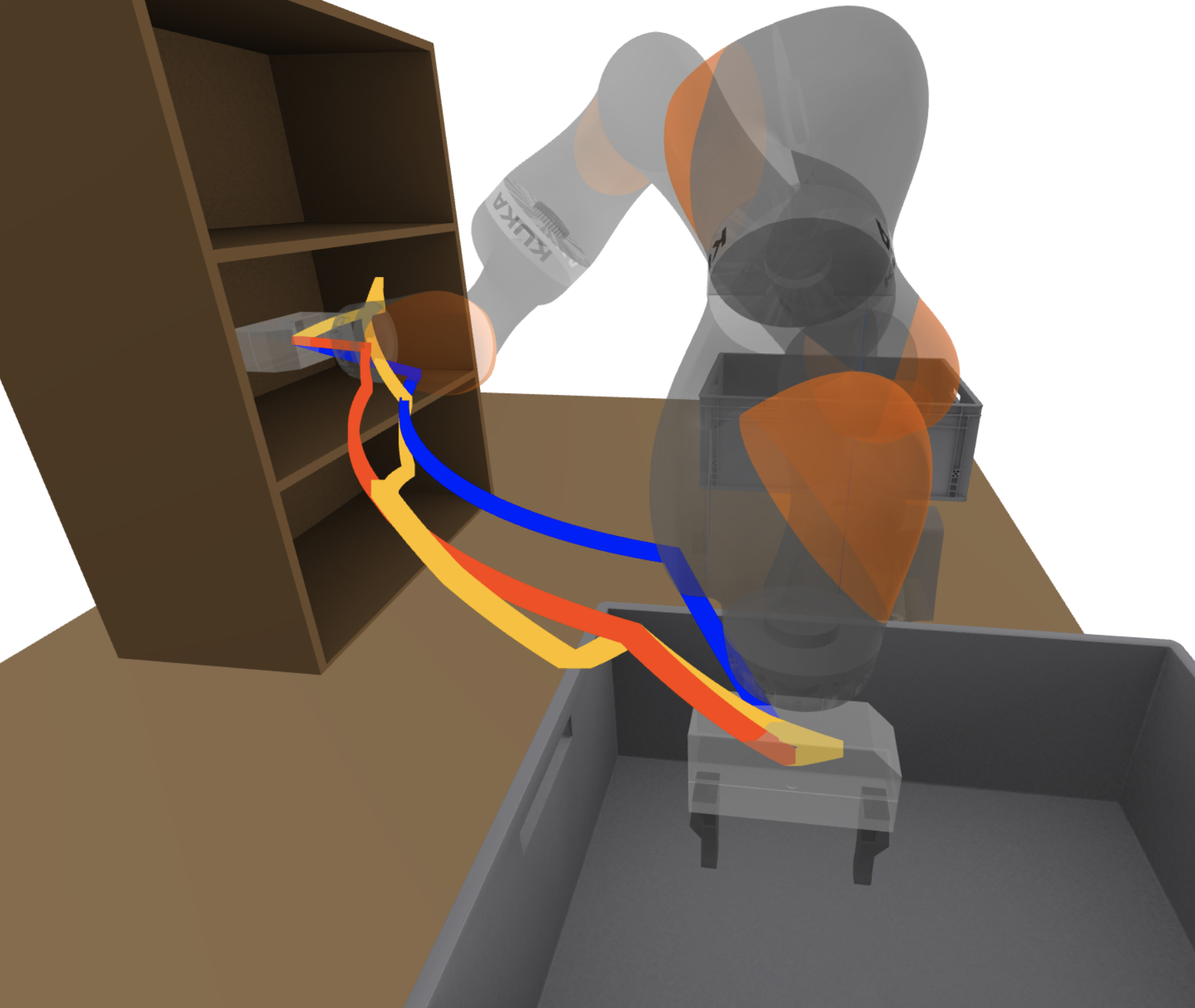}
\caption{}
\label{fig:plan_3}
\end{subfigure}
\\
\begin{subfigure}[b]{.32\textwidth}
\includegraphics[width=\textwidth]{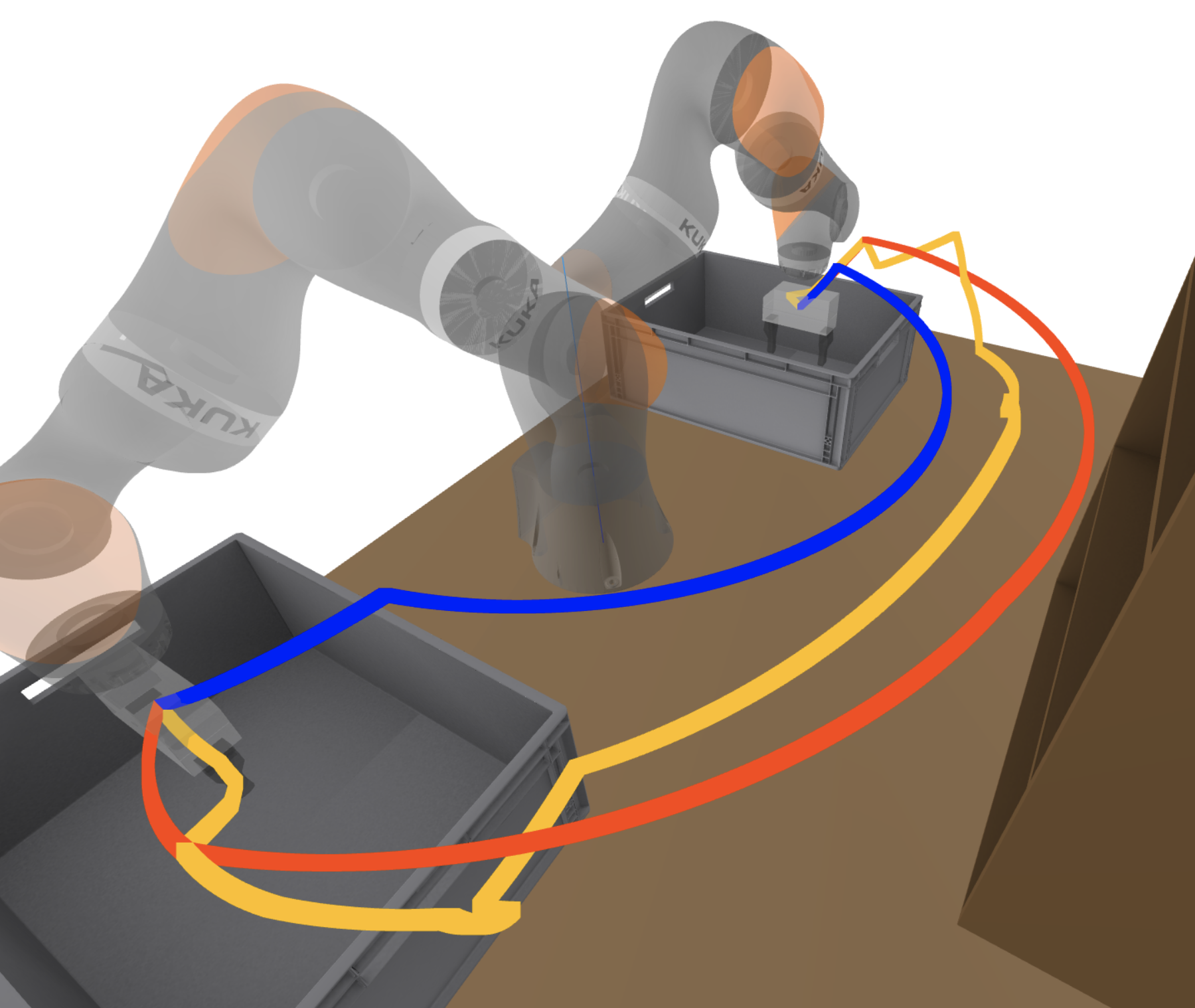}
\caption{}
\label{fig:plan_4}
\end{subfigure}
\
\begin{subfigure}[b]{.32\textwidth}
\includegraphics[width=\textwidth]{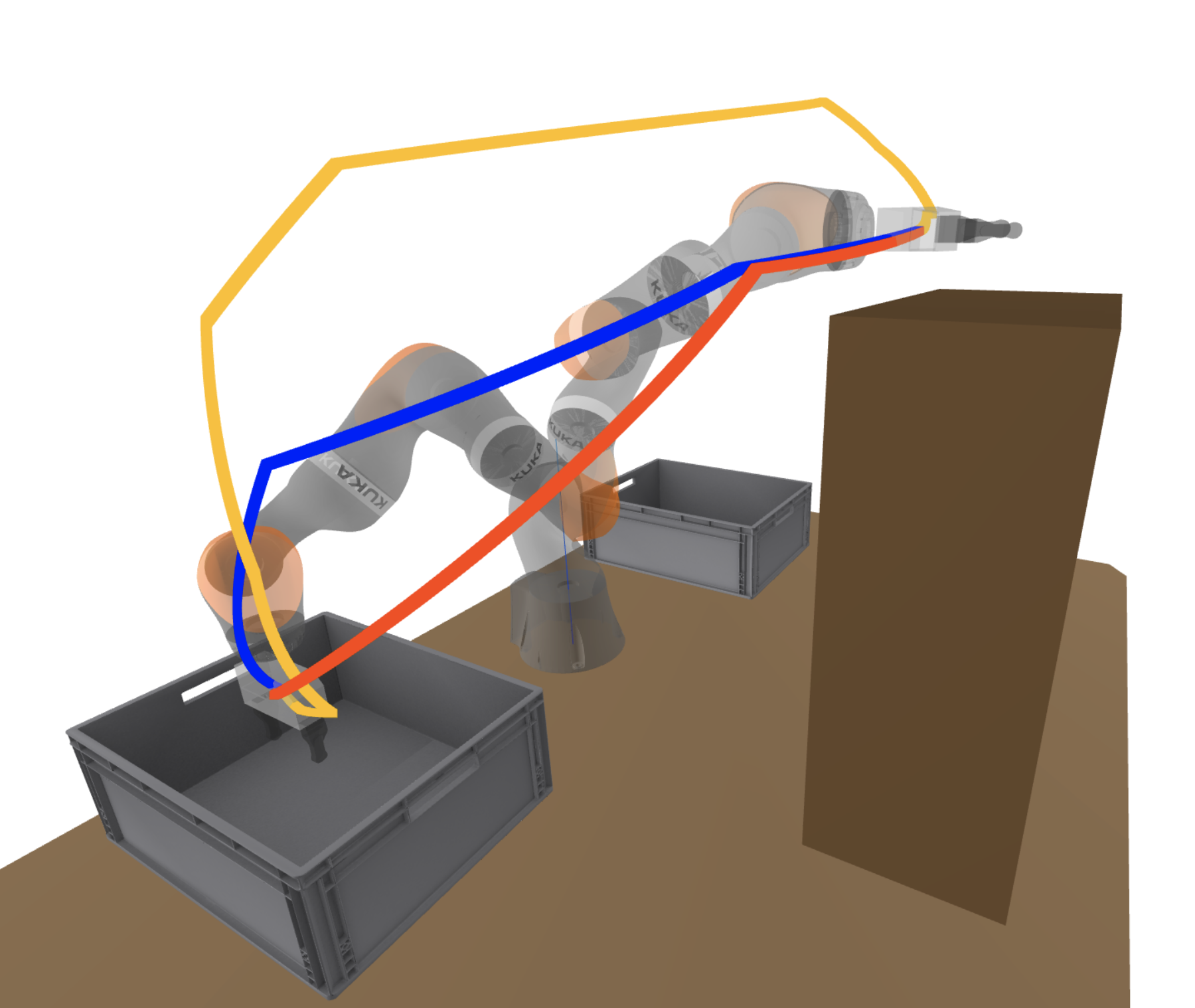}
\caption{}
\label{fig:plan_5}
\end{subfigure}
\caption{
The five motion-planning tasks for the comparison in Section~\ref{sec:comparison_with_prm}.
End-effector trajectories are depicted in blue for GCS, in yellow for the regular PRM, and in red for the PRM with short-cutting.
(a)~Task~1: from end-effector above the rack (configuration $\rho_1$) to end-effector in the upper shelf (configuration $\rho_2$).
(b)~Task~2: from upper shelf $\rho_2$ to lower shelf $\rho_3$.
(c)~Task~3: from lower shelf $\rho_3$ to left bin $\rho_4$.
(d)~Task~4: from left bin $\rho_4$ to right bin $\rho_5$.
(e)~Task~5: from right bin $\rho_5$ to above the rack $\rho_1$.
}
\label{fig:prm_tasks}
\end{figure}
\begin{figure}[t]
\centering
\begin{subfigure}[t]{0.48\textwidth}
\centering
\includegraphics[width=\textwidth]{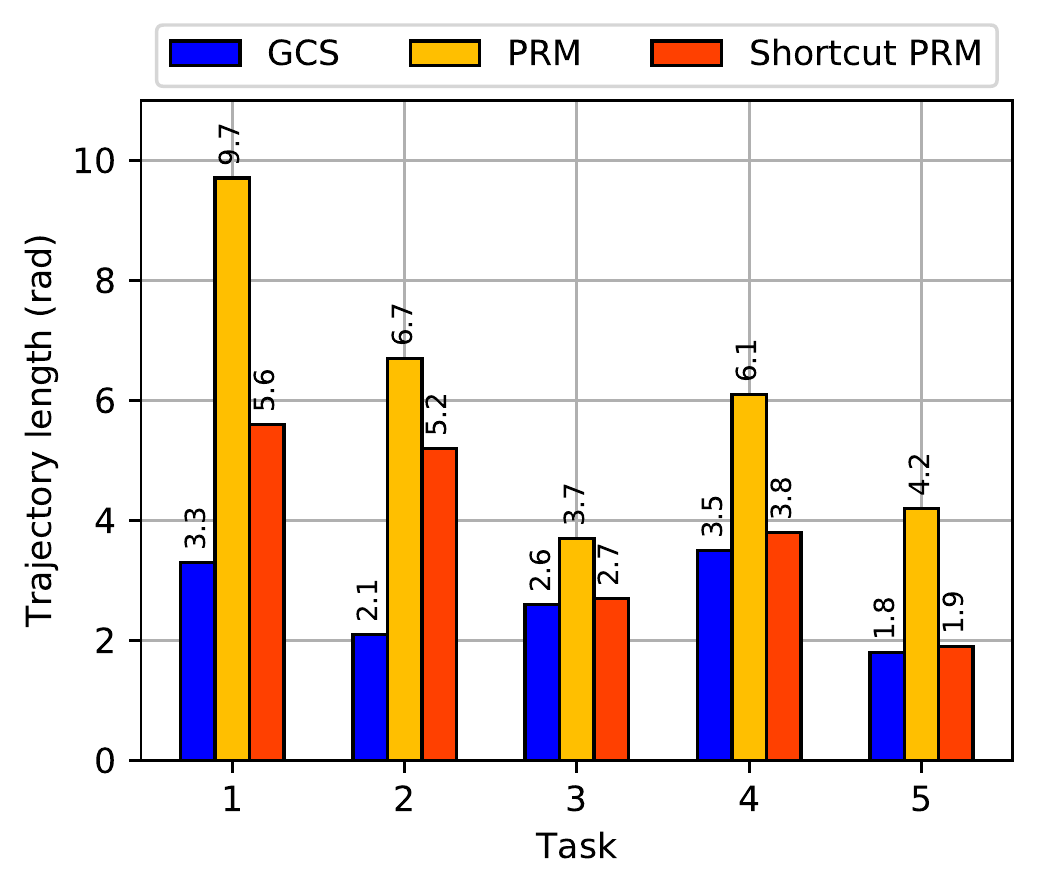}
\caption{}
\label{fig:spp_vs_prm_lengths}
\end{subfigure}
\
\begin{subfigure}[t]{0.48\textwidth}
\centering
\includegraphics[width=\textwidth]{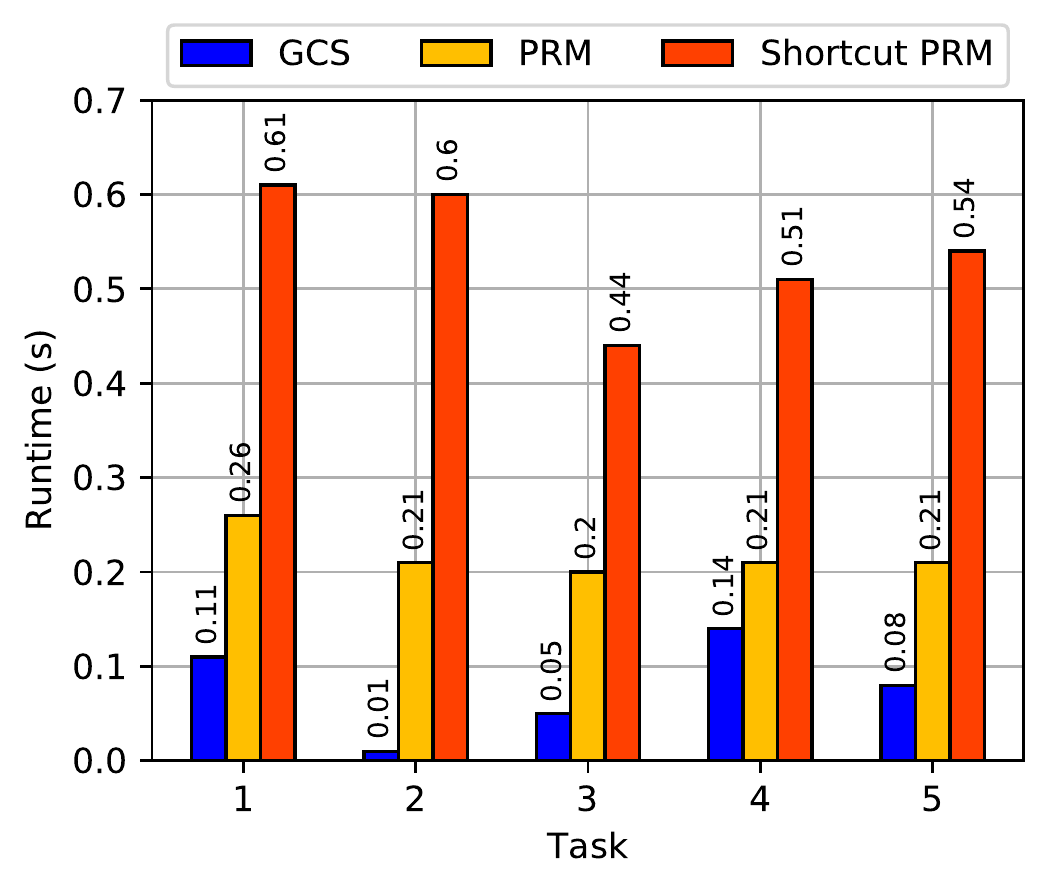}
\caption{}
\label{fig:spp_vs_prm_runtimes}
\end{subfigure}
\caption{Comparison of GCS with the PRM method and its version with short-cutting.
(a)~Length of the trajectories planned by each algorithm for the five tasks depicted in Figure~\ref{fig:prm_tasks}.
(b)~Corresponding runtimes.
GCS designs shorter trajectories than the PRM method with short-cutting, and is faster than the regular PRM.
}
\label{fig:prm_comparison}
\end{figure}

The tasks require moving the arm between five waypoint configurations $\rho_i \in \Q$, while avoiding collisions with the rack and the bins.
Each waypoint $\rho_i$ is obtained from $q_i$ by perturbing the position of  the robot end-effector as shown in Figure~\ref{fig:prm_tasks}.
We have a total of five tasks: for $i=1, \ldots, 4$, task~$i$ asks us to move the robot from $\rho_i$ to $\rho_{i+1}$; task~5 requires moving the robot from $\rho_5$ back to $\rho_1$.
The objective is to connect the start and the goal configurations with a continuous ($\eta := 0$) trajectory of minimum Euclidean length ($a:=c:=0$ and $b:=1$).
Velocity and time constraints are irrelevant given our objective.

As a visual support to the analysis, Figure~\ref{fig:prm_tasks} illustrates the trajectories of the robot end-effector generated by each planner for each task.
The blue curves correspond to GCS, the yellow to the regular PRM, and the red to the PRM with short-cutting.
Let us emphasize, though, that shorter trajectories in configuration space do not necessarily map to shorter trajectories in task space.
The actual configuration-space lengths of these trajectories are reported in Figure~\ref{fig:spp_vs_prm_lengths}, with the same color scheme.
The runtimes required by each planner can be found in Figure~\ref{fig:spp_vs_prm_runtimes}.\footnote{
The runtimes of GCS are computed by summing the times necessary for the pre-processing described in Appendix~\ref{sec:graph_preprocessing}, the solution of the convex relaxation of the SPP in GCS, and the rounding step from Section~\ref{sec:rounding_the_convex_relaxation_of_the_spp}.
}
In all the tasks, GCS designs trajectories that are shorter than both PRM methods.
Moreover, the runtimes of GCS are even smaller than the ones of the regular PRM.
The PRM with short-cutting designs higher-quality trajectories than the regular PRM, but its runtimes are significantly larger.
The pre-processing described in Appendix~\ref{sec:graph_preprocessing} is the reason why our method is extremely fast in solving task~2: in the graph $G$ in Figure~\ref{fig:graph} there is only one path that connects vertex $2$ to vertex $3$, and our pre-processing efficiently eliminates all the edges in the graph but $(2,6)$ and $(6,3)$.

In conclusion, let us mention that in all the tasks the solution we identify via rounding is the global optimum of the SPP in GCS ($\deltaopt = 0\%$).
The certified optimality gap $\deltarelax$ is 4.1\% on average, and achieves a maximum of 13.0\% in the first task.

\subsection{Coordinated Planning of Two Robot Arms}
\label{sec:coordinated_planning_of_two_robot_arms}
\begin{figure}[t]
\centering
\begin{subfigure}{0.48\textwidth}
\centering
\includegraphics[height=5cm]{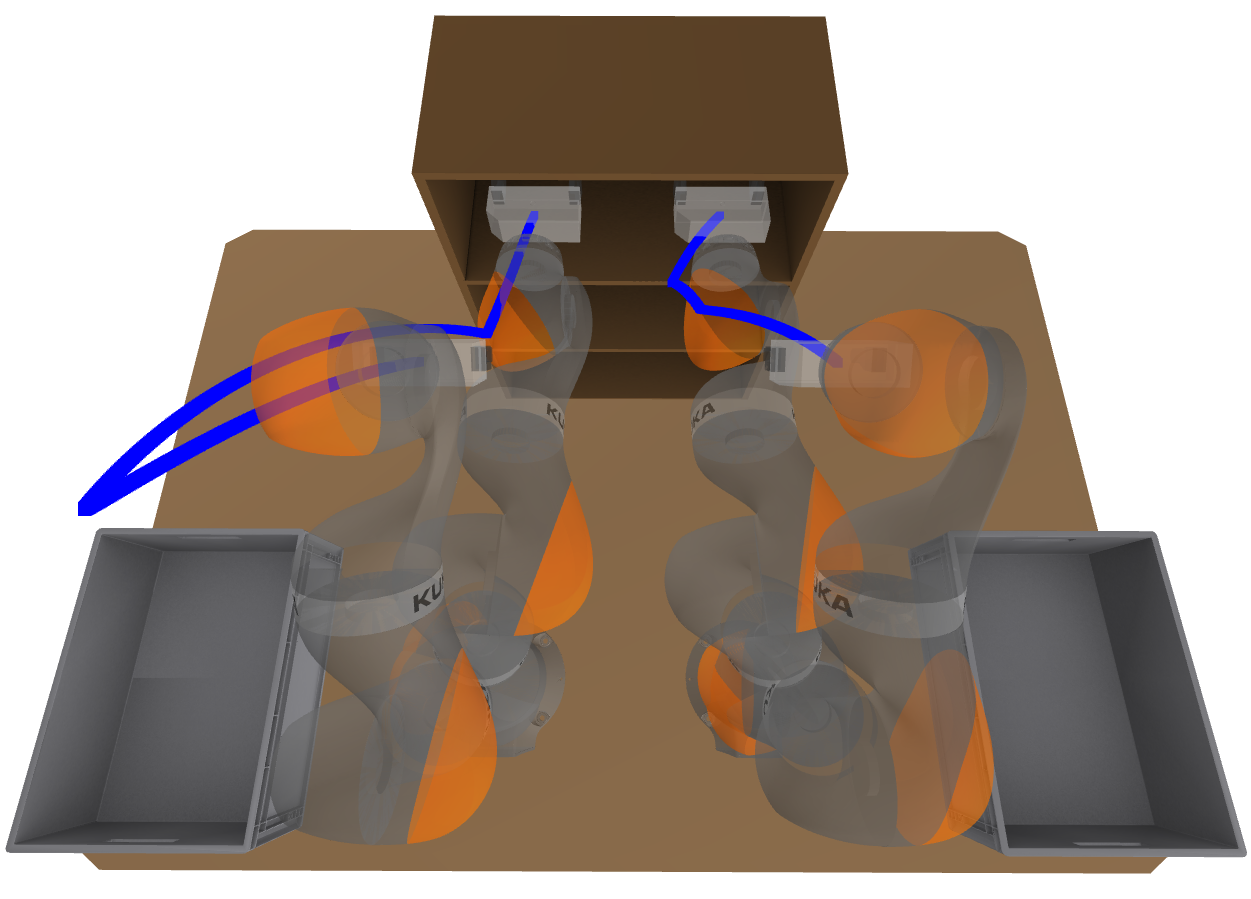}
\caption{}
\label{fig:bimaunal_1}
\end{subfigure}
\hfill
\begin{subfigure}{0.48\textwidth}
\centering
\includegraphics[height=5cm]{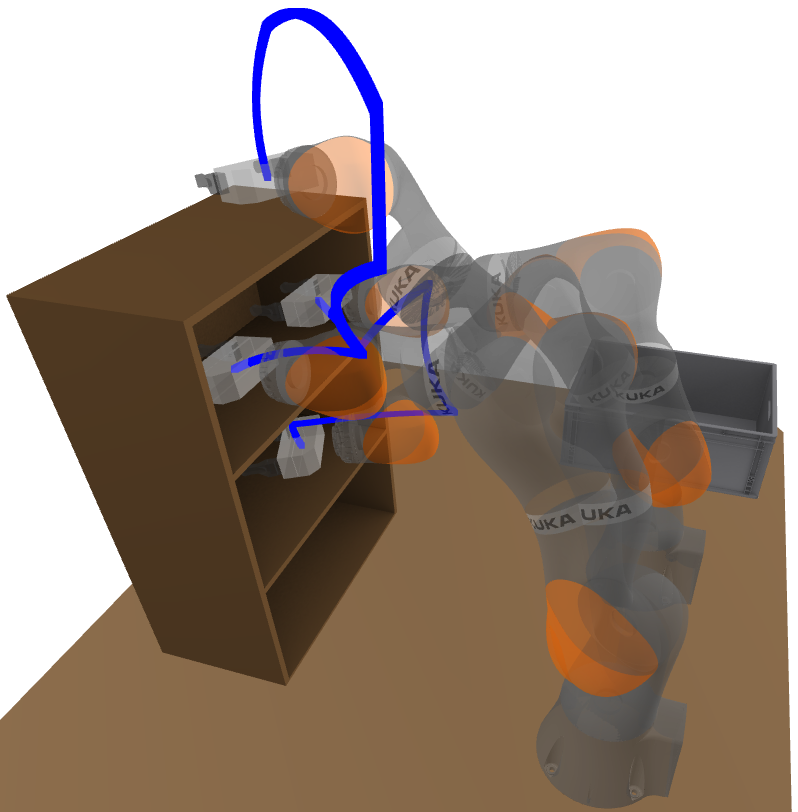}
\caption{}
\label{fig:bimaunal_2}
\end{subfigure}
\begin{subfigure}{0.48\textwidth}
\centering
\includegraphics[height=5cm]{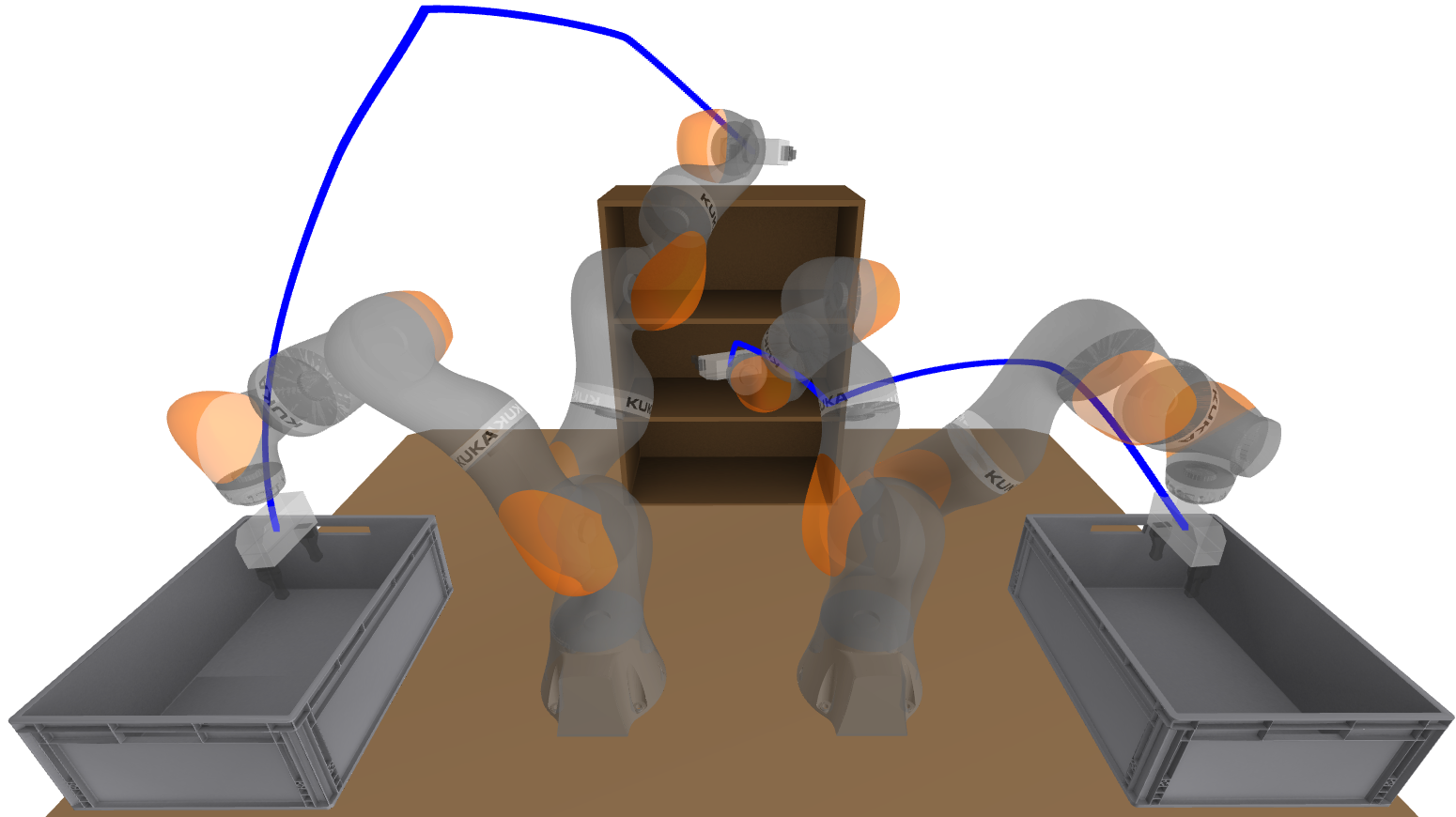}
\caption{}
\label{fig:bimaunal_3}
\end{subfigure}
\caption{
Manipulation tasks from Section~\ref{sec:coordinated_planning_of_two_robot_arms}.
End-effector trajectories are in blue.
(a)~Task~1: arms from neutral pose to top shelf.
(b)~Task~2: from top shelf to configuration with crossed arms.
(c)~Task~3: from crossed arms to lateral bins.
Despite the fourteen-dimensional configuration space, the potential collisions between the arms, and the confined environment, GCS can reliably solve the three tasks in a few seconds via convex optimization.
}
\label{fig:bimanual_trajectory}
\end{figure}

In the previous subsection we have compared GCS to widely-used PRM methods, choosing a robotic arm with $n=7$ degrees of freedom because sampling-based algorithms perform poorly in higher dimensions.
Here we demonstrate that GCS can tackle planning problems in much higher-dimensional spaces.
To this end, we consider the dual-arm manipulator shown in Figure~\ref{fig:bimanual_trajectory}, composed of two KUKA LBR iiwa with seven degrees of freedom each, yielding an overall configuration space $\Q$ of $n=14$ dimensions.
The environment is the same as in the previous subsection, but this time, besides the collisions with the rack and the bins, GCS must also prevent collisions between the arms themselves.

To decompose the configuration space we proceed as in Section~\ref{sec:comparison_with_prm}.
This time we use a total of 22 seed poses, chosen to approximately cover the workspace around the rack and the bins, as well as the rest of the free space.
Also in this case the seeds are produced manually, using inverse kinematics and with the visual support provided by the connectivity of the graph $G$.
We analyze three tasks.
In the first task, illustrated in Figure~\ref{fig:bimaunal_1}, the arms start in a neutral position and both reach into the top shelf.
Task~2, in Figure~\ref{fig:bimaunal_2}, asks the arms to cross: the left arm reaches above the rack on the right, and the right arm moves to the left of the bottom shelf.
Finally, in Figure~\ref{fig:bimaunal_3}, task~3 requires the two arms to reach inside the bins.
To make the problem even more challenging, this time we do not limit ourselves to the design of purely-geometric shortest curves as in Section~\ref{sec:comparison_with_prm}, but we plan continuously differentiable ($\eta := 1$) trajectories of degree $d:=3$.
The weights in the objective~\eqref{eq:statement_objective} are set to $a:=b:=1$ and $c:=0$.
The constraint set $\D$ in~\eqref{eq:statement_derivative_constraints} ensures that the joint velocities are no greater than 60\% of the robot velocity limits.
The duration bounds $\Tmin$ and $\Tmax$ are set so that they do not affect the optimal trajectory, while the boundary values of the velocity are zero ($\dot q_0 := \dot q_T := 0$).
As described in Section~\ref{sec:additional_differential_costs_and_constraints}, we penalize accelerations via a cost term of the form~\eqref{eq:regularizer}, with weight $\varepsilon = 10^{-3}$.
With the same goal, we set $\dot h_\mathrm{min} := 10^{-3}$.

The trajectories synthesized by GCS for each of the three tasks are represented in Figure~\ref{fig:bimanual_trajectory}, with the curves swept by the end-effectors depicted in blue.
The optimality gaps $\deltarelax$ certified by GCS for the three tasks are $3.3\%$, $2.0\%$, and $0.6\%$.
Running a mixed-integer solver, we verify that the first two trajectories are, in fact, globally optimal, while the last trajectory has an optimality gap of only $\deltaopt = 0.3\%$.
As in Section~\ref{sec:statistical_analysis_quadrotor_flying_through_buildings}, to circumvent numerical issues, we set the MOSEK option \texttt{MSK\_IPAR\_INTPNT\_SOLVE\_FORM = 1} in the solution of the convex relaxations.
This leads to the following computation times for the three tasks at hand: $4.0$~s, $8.4$~s, and $12.9$~s.
As already mentioned, we are confident that a tailored pre-solve stage can drastically decrease these runtimes.

\section{Discussion}
\label{sec:discussion}
On the one hand, transcribing the motion-planning problem as an SPP in GCS allows us to use efficient convex optimization to design trajectories around obstacles.
On the other hand, our convexity requirements restrict the class of planning problems we can tackle, and limit the families of trajectories we can parameterize.
In this section we comment on the strengths and the limitations of our approach, and we illustrate the pros and the cons of GCS over existing planning algorithms.

\subsection{Additional Costs and Constraints}
\label{sec:additional_costs_and_constraints}

Besides the derivative penalties discussed in Section~\ref{sec:additional_differential_costs_and_constraints}, there are many additional costs and constraints that our problem statement~\eqref{eq:statement} does not feature but that are relevant in a variety of practical applications.
Minimum-distance and minimum-time objectives might lead to unsafe robot trajectories, that do not avoid obstacles with sufficient clearance.
A practical workaround in these cases is to discourage the control points of our trajectories to get too close to certain boundaries of the safe regions $\Q_i$.
This can be achieved through convex barrier penalties, that are easily included among the edge costs in Section~\ref{sec:the_edge_lengths_elle}.
Equality constraints that couple the trajectory $q$ to its time derivatives could be used to enforce continuous-time dynamics.
However, our choice of optimizing over the shape $r$ and the timing $h$ of the trajectory jointly makes these constraints nonconvex, even for a linear control system.
Similarly, the nonlinearity of the kinematics of a robot manipulator makes task-space constraints not directly suitable for our framework.
To cope with these nonconvexities, in some applications, it may be practical to post-process the output of GCS with a local nonconvex optimizer.

\subsection{Comparison with Existing Mixed-Integer Planners}

GCS has three main advantages over existing MICP algorithms for solving problems of the form~\eqref{eq:statement}:
\begin{enumerate}
\item
The tightness of the convex relaxation of our MICPs, demonstrated empirically in the numerical results in Section~\ref{sec:numerical_results}.
\item
The reduced number of binary decision variables in our programs, illustrated in the maze example from Section~\ref{sec:motion_planning_in_a_maze}.
\item
The simplicity of the class of optimization problems that our method leads to, discussed in Section~\ref{sec:class_of_optimization_problems}
\end{enumerate}
The first and the second are achieved by leveraging the optimization framework from~\cite{marcucci2021shortest}.
The third is partly due to the first (since it is the tightness our MICP formulations that allows us to tackle the motion planning problem as a single convex program, plus rounding), but it is also due to the parameterization of trajectories as B\'{e}zier curves.

In Section~\ref{sec:the_convex_sets_Xv}, we have leveraged the properties of B\'{e}zier curves to enforce infinite families of constraints through a finite number of conditions.
For example, in~\eqref{eq:Xi_containment}, we have transcribed the safety requirement $r_i(s) \in \Q_i$ for all $s \in [0,1]$ as a constraint $r_{i,k} \in \Q_i$ per control point $k=0, \ldots, d$.
The MICP planner from~\cite{deits2015efficient} achieves the same result by using Sums-Of-Squares (SOS) polynomials~\cite{parrilo2000structured}, and semidefinite programming.
These approaches are interchangeable and lead to a tradeoff: B\'{e}zier curves yield simpler constraints, SOS polynomials parameterize a richer class of trajectories.\footnote{
Asking a univariate polynomial to be nonnegative by parameterizing it as a B\'{e}zier curve with nonnegative control points is more stringent than asking it to be SOS (which, in the univariate case, is equivalent to nonnegativity).
}
In the numerical examples analyzed in this paper, we have found this gap to be relatively narrow, and we have then prioritized simpler optimization problems.

Finally, it is worth mentioning that the problem formulation from~\cite{deits2015efficient} features costs and constraints on time derivatives of the trajectory $q$ of any order.
These, however, are handled by fixing the duration of each trajectory segment beforehand.
A similar result could be achieved with GCS by fixing the time that can be spent in each safe set $\Q_i$.

\subsection{Comparison with Sampling-Based Algorithms}

As discussed in Section~\ref{sec:comparison_with_prm}, among many sampling-based planners, PRM is the natural comparison for GCS.
In fact, GCS can be thought of as a generalization of the PRM method, where each collision-free sample is expanded to a collision-free convex region, that is inflated as much as the obstacles allow; reducing in this way a dense roadmap to a compact GCS.
In Sections~\ref{sec:comparison_with_prm} and~\ref{sec:coordinated_planning_of_two_robot_arms}, we have shown that GCS can outperform PRM in terms of: runtimes, quality of the designed trajectories, scalability with the dimensionality $n$ of the configuration space $\Q$, and variety of objective functions and trajectory constraints.
In addition, because of the parallel above, it is reasonable to imagine that  many of the techniques developed for PRM to handle, e.g., changes in the environments~\cite{jaillet2004prm,van2006anytime} can be translated to GCS with relatively low effort.

One of the main reasons why sampling-based methods are widely used in academia and industry is their simplicity.
Conversely, the implementation of GCS is very involved and requires familiarity with convex-optimization techniques.
Nonetheless, we believe that the framework from~\cite{marcucci2021shortest} lends itself to an intuitive mathematical abstraction, and that the programming interface of GCS can be made very easy to use.
We have provided a mature implementation of the techniques from~\cite{marcucci2021shortest} within the open-source software Drake~\cite{tedrake2019drake}, and we have developed a simple GCS interface at~\url{https://github.com/mpetersen94/gcs}.

\subsection{Comparison with Direct Trajectory Optimization}

Direct-trajectory-optimization methods transcribe the motion-planning problem into a nonconvex optimization~\cite{diehl2006fast}, and can virtually include any sort of cost terms and constraints, including dynamic and task-space constraints.
In practice, however, these nonconvex programs can only be tackled with local-optimization algorithms that are slow and unreliable.
GCS is different in spirit, as we prioritize low runtimes and the completeness of the planning algorithm over the modelling power.

\section{Conclusions and Future Works}
In this paper we have introduced GCS: an algorithm based on convex optimization for efficient collision-free motion planning.
GCS leverages the framework presented in~\cite{marcucci2021shortest} to design a very tight and lightweight convex relaxation of the planning problem.
This convex optimization (typically an SOCP) is quickly solved using commonly-available software, and a cheap randomized rounding of its solution is almost always sufficient to identify a globally-optimal trajectory.
We have demonstrated GCS on a variety of scenarios: an intricate maze, a quadrotor flying through buildings, and a manipulation task in a fourteen-dimensional configuration space.
Furthermore, we have compared GCS to widely-used PRM methods, showing that our method can find higher-quality trajectories in less time.

This paper presents the first version of a new algorithm, which already compares favorably with widely-used planners that have been optimized over decades.
The runtimes of GCS can be drastically reduced (we are currently developing a customized solver for these convex optimizations).
We are also highly optimistic that the class of cost functions and constraints that we can handle will expand considerably in the future.
In particular, we imagine incorporating task-space constraints, tight penalties on the higher derivatives of the trajectory, as well as dynamic constraints arising from input limits.
Furthermore, we wish to extend GCS to problems involving contacts between the robot and the environment.
We believe that our planner demonstrates the value of formulating problems as SPPs in GCS, and it can already find multiple real-world applications.

\appendix

\section{Further Details on the Implementation of GCS}
\label{sec:further_details_on_the_implementation_of_gcs}
In this appendix we illustrate two techniques that we employed in the numerical results in Section~\ref{sec:numerical_results} to tighten and compress the convex relaxations of our planning problems.

\subsection{Two-Cycle-Elimination Constraints}
\label{sec:two_cycle_elimination_constraints}

The graph $G$ constructed in Section~\ref{sec:the_graph_G} connects each pair $\Q_i$ and $\Q_j$ of overlapping safe regions with a two-cycle: $e:=(i,j)$ and $f:=(j,i)$.
Since by traversing both the edges $e$ and $f$ we would visit vertex $i$ twice, and this is not allowed by the definition of a path $p$, at least one of these edges must be excluded from the shortest path.
In other words, the indicator variables $\varphi_e$ and $\varphi_f$ cannot be both equal to one.
This observation can be used to tighten our convex relaxations, and speed up our planner.

More precisely, for each pair of overlapping regions, we can write the linear constraints
\begin{align}
\label{eq:two_cycle}
\varphi_e + \varphi_f \leq \varphi_i
\quad \text{and}\quad
\varphi_e + \varphi_f \leq \varphi_j,
\end{align}
where $\varphi_i$ and $\varphi_j$ represent the total probability flows traversing vertices $i$ and $j$, respectively.
(Note that, since the total flow through a vertex is at most one, these inequalities imply the looser condition $\varphi_e + \varphi_f \leq 1$.)
Furthermore, by applying Lemma~1(b) from~\cite{marcucci2021shortest}, the two inequalities in~\eqref{eq:two_cycle} can be translated into a pair of convex constraints that tighten the coupling between the flow variables $\varphi_e$ and the continuous variables $x_v$ in our convex programs.
The number of these constraints is linear in the size $|\E|$ of the edge set, and they can substantially increase the tightness of the convex relaxations of our planning problems.
They are enforced in all the numerical results presented in Section~\ref{sec:numerical_results}.

\subsection{Graph Pre-Processing}
\label{sec:graph_preprocessing}

The constraints described in Appendix~\ref{sec:two_cycle_elimination_constraints} represent only one of the multiple ways in which we can leverage the knowledge that a path $p$ is allowed to visit a vertex at most once.
For example, consider the graph in Figure~\ref{fig:graph} and task~2 from Section~\ref{sec:comparison_with_prm} of moving the robot arm between the configurations $\rho_2 \in \Q_2$ and $\rho_3 \in \Q_3$.
In this case, after connecting the source $\sigma$ to vertex $2$ and vertex $3$ to the target $\tau$, we get a graph that admits a single $\sigma$-$\tau$ path: $p := (\sigma, 2, 6, 3, \tau)$.
Therefore, in this particular case, a pre-processing stage capable of making such an inference would reduce our planning problem to a tiny convex program (exactly).

In general, making the inference just described exactly is infeasible; however, in many practical scenarios, a cheap approximate pre-processing can eliminate most of the redundancies in our graphs $G$.
More precisely, checking if an edge $e :=(u,v)$ can be traversed by a $\sigma$-$\tau$ path is equivalent to solving a vertex-disjoint-paths problem.
This problem asks to identify a path $p_1$ from $\sigma$ to $u$ and a path $p_2$ from $v$ to $\tau$ such that the overall path $p:= (p_1, p_2)$ is a valid path from $\sigma$ to $\tau$.
In other words, the two subpaths $p_1$ and $p_2$ are not allowed to share any vertex.
This problem is NP-complete~\cite[Section~70.5]{schrijver2003combinatorial}, therefore it would not make sense to solve it exactly as a pre-processing for our planner.
Nevertheless, the vertex-disjoint-paths problem admits a natural LP relaxation as a fractional multiflow problem~\cite[Section~70.1]{schrijver2003combinatorial} that can be solved very quickly, and can be used as a very-effective sufficient condition to check if an edge is redundant.

We have found this pre-processing to be particularly useful when the graph $G$ is sparse and has small size, and the convex sets $\X_v$ associated to its vertices live in high dimensions.
In these cases, the multiflow LPs (which can be tackled in parallel) are solved extremely fast and they can drastically compress and tighten our convex optimizations.
We have employed this pre-processing strategy in the numerical examples from Sections~\ref{sec:statistical_analysis_quadrotor_flying_through_buildings},~\ref{sec:comparison_with_prm}, and~\ref{sec:coordinated_planning_of_two_robot_arms}: the runtimes of GCS reported in these sections include the time necessary for pre-processing.

\section{Random Environment Generation for the Quadrotor Example}
\label{sec:random_environment_generation}
In this appendix we briefly describe the algorithm we employed for the generation of the random buildings in Section~\ref{sec:statistical_analysis_quadrotor_flying_through_buildings}.

The buildings are constructed over a five-by-five grid, where each cell has sides of length $5$.
The nine cells at the center of the grid are occupied either by a room, a tree, or obstacle-free grass.
The sixteen cells at the boundary of the grid are always occupied by grass.
For all the environments, the brown start block is in the cell $(1,1)$, while the green goal block is in the cell $(4,3)$ (see Figure~\ref{fig:quadrotor}).
To assemble a building we start from the goal cell, which we always require to be a room.
Then we mark each adjacent cell either as inside or outside the building, and we repeat this process until the nine inner cells are occupied.
For the cells that are marked as outside the building, we decide at random whether to grow a tree or not.
Walls divide the rooms from the outside, and are built with either a doorway, a window, two windows, or no openings at all.
Walls are also used to divide the rooms; in this case we randomly select a doorway, a vertical half wall, a horizontal half wall, or no wall.
The positions of the trees are also drawn at random, while their sizes are taken to be constant.

Given that the walls and the trees have polygonal shape, the decomposition of the configuration space $\Q$ can be done exactly.
Specifically, we pair rooms or cells that are occupied by grass with a single box $\Q_i$ of free space, while the space around a tree is decomposed using four non-overlapping boxes.
Suitable box-shaped regions are added for each (inner or outer) wall that contains one or more openings.
Finally, the safe regions $\Q_i$ are adequately shrunk to take into account the collision geometry of the quadrotor, which is taken to be a sphere of radius $0.2$.

\section{Implementation of the PRM Planner}
\label{sec:implementation_of_the_prm}
In this appendix we report the main implementation details for the PRM and the short-cutting algorithm used in the comparison in Section~\ref{sec:comparison_with_prm}.

We construct the PRM using the implementation~\href{https://github.com/calderpg/common_robotics_utilities/blob/baf66f7c31d351d4f9f1e61508b6242ae91d7fe5/include/common_robotics_utilities/simple_prm_planner.hpp}{\texttt{simple\_prm\_planner.hpp}} from the library~\cite{cru}.
Trying to construct a roadmap just by sampling random robot poses turns out to be infeasible for the application  in Section~\ref{sec:comparison_with_prm}.
In fact, sampling a robot pose $q \in \R^7$ for which the end effector is, e.g., inside one of the shelves in Figure~\ref{fig:kuka_graph} is extremely unlikely: after $3 \cdot 10^5$ samples, and $90$ hours of computations, we did not find any such point.
As a result, we construct the roadmap in two steps.
In the first step, we connect the seed poses $\{q_i\}_{i=1}^8$ from Figure~\ref{fig:kuka_graph} using a collection of bidirectional Rapidly-exploring Random Trees (RRTs) (\href{https://github.com/calderpg/common_robotics_utilities/blob/baf66f7c31d351d4f9f1e61508b6242ae91d7fe5/include/common_robotics_utilities/simple_rrt_planner.hpp}{\texttt{simple\_rrt\_planner.hpp}} from~\cite{cru}).
The role of these trees is to form a skeleton for the PRM, and, to keep this skeleton reasonably compact, we mimic the connectivity of our graph $G$ in Figure~\ref{sec:the_graph_G}.
In particular, we connect via RRT only the pairs of seed poses $q_i$ and $q_j$ for which the vertices $i$ and $j$ are connected in $G$.
This process gives us $12$ trees, with a total of approximately $2,300$ nodes.
In the second step, we fill the rest of the space according to the standard PRM algorithm.
We stop the sampling when we reach a total of $15 \cdot 10^3$ nodes in the PRM, included the ones from the RRTs.
(In our experience, a larger number of PRM samples would have led to an increase in the runtimes without sensibly improving the quality of the designed trajectories.)
During this construction, the collision checks are handled by Drake~\cite{tedrake2019drake}.
With this setup, generating the RRTs took a total of 60~seconds, while the remaining PRM samples required 15~minutes.
For the short-cutting algorithm we use the implementation in~\href{https://github.com/calderpg/common_robotics_utilities/blob/baf66f7c31d351d4f9f1e61508b6242ae91d7fe5/include/common_robotics_utilities/path_processing.hpp}{\texttt{path\_processing.hpp}} from~\cite{cru}.

The numerical parameters we use for the RRT, the PRM, and the short-cutting algorithm are chosen to optimize the tradeoff between the quality of the designed paths and the overall computation times.


\section*{Acknowledgements}
We would like to thank Greg Izatt for providing us with the environment generator used to benchmark our algorithm in Section~\ref{sec:statistical_analysis_quadrotor_flying_through_buildings}, and for his precious assistance with the visualization of the motion plans in Sections~\ref{sec:comparison_with_prm} and~\ref{sec:coordinated_planning_of_two_robot_arms}.
We are grateful to Andres Valenzuela for his great help in the initial phases of this project, and to Pablo A. Parrilo, Jack Umenberger, and Calder Phillips-Grafflin for the many insightful suggestions and discussions.

This material is based upon work supported by (in alphabetical ordered):
Amazon.com, PO No. 2D-06310236;
Department of Defense (DoD) through the National Defense Science \& Engineering Graduate (NDSEG) Fellowship Program;
Lincoln Laboratory/Air Force, PO No. 7000470769;
National Science Foundation, Award No. EFMA-1830901;
Office of Naval Research, Award No. N00014-18-1-2210; and
Under Secretary of Defense for Research and Engineering, Air Force Contract No. FA8702-15-D-0001.
Any opinions, findings, conclusions, or recommendations expressed in this material are those of the authors and do not necessarily reflect the views of the funding agencies.

\bibliographystyle{plainurl}
\bibliography{bibliography}

\end{document}